\documentclass{article}
\usepackage[T1]{fontenc}
\usepackage[utf8]{inputenc}
\usepackage{xcolor}

\usepackage{latexsym}
\usepackage{microtype}
\usepackage{inconsolata}
\usepackage{graphicx}
\usepackage{adjustbox}
\usepackage{booktabs}
\usepackage{multirow}
\usepackage{enumitem}
\usepackage{xspace}
\usepackage{tabularx}
\usepackage{array}
\usepackage{algorithm}
\usepackage{algpseudocode}
\usepackage{makecell}
\usepackage{capt-of}
\usepackage[breakable]{tcolorbox}
\usepackage{placeins}
\usepackage{alltt}
\usepackage{arxiv}

\newcolumntype{Y}{>{\raggedright\arraybackslash}X}
\newcolumntype{P}[1]{>{\raggedright\arraybackslash}p{#1}}

\newcommand{\method}{\textsc{MARS}\xspace}
\newcommand{\qtm}{\textsc{QTM}\xspace}
\newcommand{\realm}{\textsc{RealMem}\xspace}
\newcommand{\mc}{\textsc{MultiChallenge}\xspace}

\title{\method: Memory-Aware Routing for Session-Level LLM Assistants}

\author{
    Minghan Wang\textsuperscript{1},
    Jinhang Zuo\textsuperscript{2},
    Zhongxiang Dai\textsuperscript{3},\\
    Zhiyong Wang\textsuperscript{4},
    Fang Kong\textsuperscript{1}
}

\affiliation{
    \textsuperscript{1}Southern University of Science and Technology\\
    \textsuperscript{2}City University of Hong Kong\\
    \textsuperscript{3}The Chinese University of Hong Kong, Shenzhen\\
    \textsuperscript{4}Harbin Institute of Technology (Shenzhen)
}

\date{}

\hypersetup{
    pdftitle={MARS: Memory-Aware Routing for Session-Level LLM Assistants}
}

\begin{document}

\maketitle

\begin{abstract}
Long-term LLM assistants must repeatedly decide what prior context to expose and which model should answer as an interaction evolves.
Existing memory systems and model routers usually make these decisions separately, although memory construction changes the evidence, distractors, and difficulty of the resulting input.
We introduce \method, a memory-aware routing framework that uses session state as a shared control interface between memory construction and model selection.
A Query and Topic Manager (QTM) estimates compact decision-relevant state that guides a fixed state-conditioned memory policy and an online router, enabling each turn to receive an appropriate memory view and a suitable model under a quality--memory--cost trade-off.
We evaluate \method on \realm and \mc with 16 system configurations each, covering long-history memory-driven interaction and bounded sessions with structured context interference.
On \realm, \method preserves 98.2\% of full-context GPT-4o's QA while using only 14.5\% of its generation API cost, including both generation and controller calls for \method, and achieves the strongest measured memory-recall and retrieval scores among compared systems.
On \mc, \method trails the strongest static system overall but approaches its performance at lower recorded cost on a mechanism-focused subset.
Together, these results support a favorable quality--memory--cost trade-off for long-history, memory-dependent interaction.
\end{abstract}
\section{Introduction}
\label{sec:intro}

Modern LLM assistants are no longer used only for isolated questions; they increasingly support ongoing work across many turns and sessions. In long-term use, interactions can span many turns and recur across tasks, a setting highlighted by recent work on long-term conversational and memory-driven agents~\citep{maharana-etal-2024-evaluating,park2023generative,packer2023memgpt,bian2026realmem}. Users may resume unfinished artifacts, request revisions, refer to earlier constraints, or ask for final answers that depend on decisions made several turns or sessions earlier. Before answering such turns, an assistant must decide both what prior context to include and which model should answer.
This creates a quality--memory--cost trade-off: richer memory can provide necessary evidence but also increases context length and distractors, while stronger models can better interpret complex memory views but are more expensive.

\begin{figure}[htbp]
\centering
\includegraphics[width=0.97\textwidth]{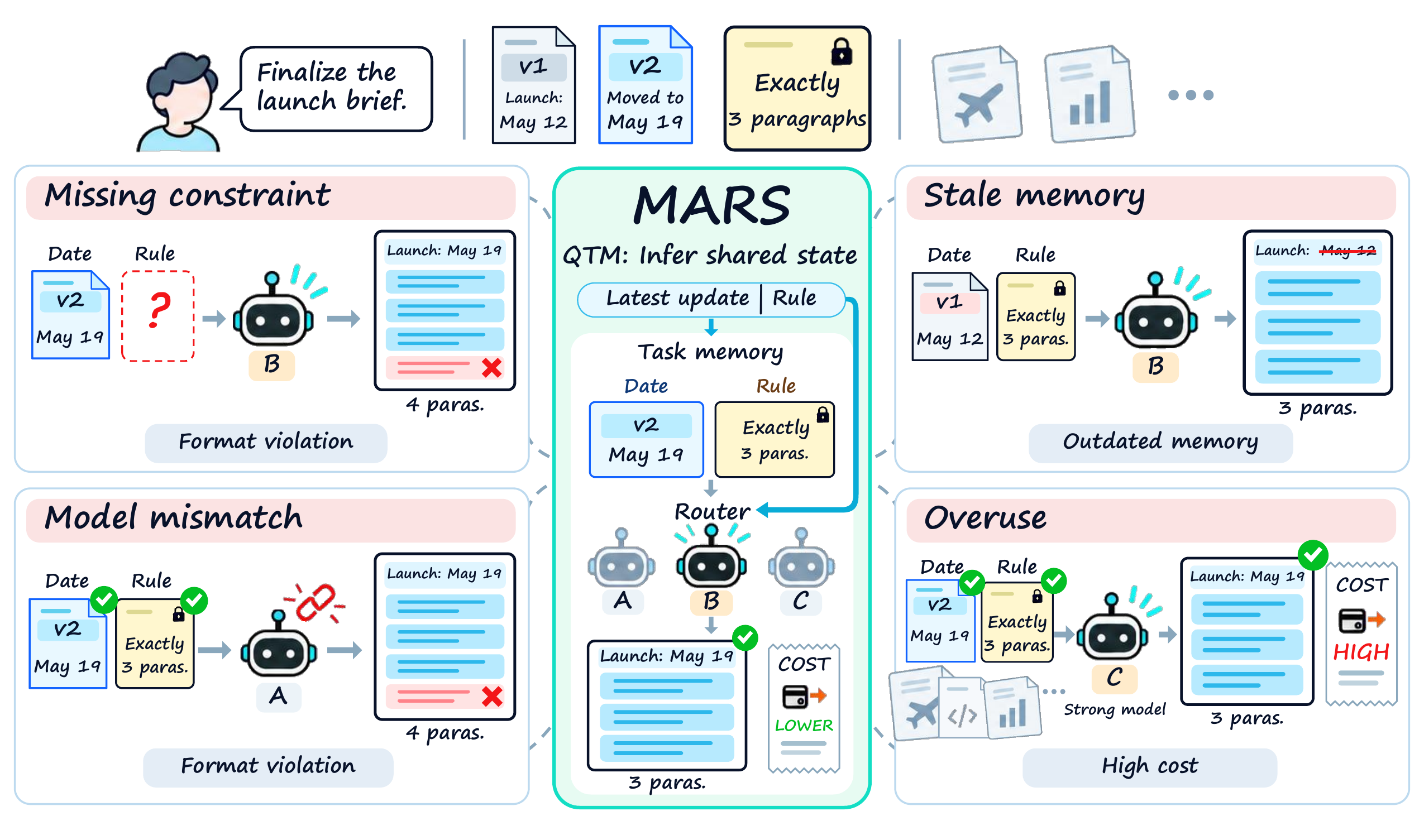}
\caption{
Illustrative motivation for memory-aware routing in a multi-turn session.
\method uses a shared QTM-derived session state to coordinate task-relevant memory construction and online model routing.
The surrounding panels illustrate missing constraints, stale memory, model mismatch, and costly overuse.
A--C denote candidate models; outputs are schematic and costs qualitative, with ``lower'' relative to the overuse configuration.
}
\label{fig:intro-example}
\end{figure}

Existing systems usually handle these decisions separately.
Memory systems store, update, and retrieve prior information~\citep{lewis2020retrieval,packer2023memgpt,chhikara2025mem0,xu2025amem}, while model routers select a generator from a candidate pool to trade off quality and cost~\citep{chen2023frugalgpt,hu2024routerbench,ding2024hybrid,ong2024routellm,nguyen2024metallm,poon2025online,wang2025mixllm}.
This separation overlooks a key dependency: memory construction determines what evidence, constraints, and distractors enter the prompt, while routing determines whether the selected model can use that context reliably and cost-effectively. 

This dependency makes memory construction and model routing inherently coupled.
As Figure~\ref{fig:intro-example} illustrates, finalizing a launch brief requires using the updated launch date (May~19, superseding May~12) while preserving an earlier constraint of exactly three paragraphs.
Omitting the constraint or exposing the obsolete date can cause errors, while an unsuitable model may fail even when the relevant memory is present.
Exposing all history can add distractors and increase context difficulty~\citep{liu2024lost,bai2023longbench}, and always invoking a stronger model can incur unnecessary cost.
The appropriate model depends on the selected memory view rather than the raw query alone.
The key challenge is therefore to coordinate what history becomes input and which model can use that context reliably and cost-effectively.

Motivated by this coupling, we study \emph{session-level adaptive memory routing}: coordinating memory exposure and model selection throughout a session under a quality--memory--cost trade-off.
We formulate this as a sequential decision problem: the system estimates session state, constructs a candidate memory view, selects a model, packs the view for that model, generates a response, and updates its routing preference from turn-level feedback.
Because the memory view changes the evidence, distractors, context length, and reasoning burden faced by the selected model, the system needs a shared session-state interface for both memory construction and model routing.

We introduce \textbf{\method}, a memory-aware routing framework for session-level LLM assistants.
\method uses a Query and Topic Manager (\qtm) to infer structured session state from the query, active thread, recent turns, and memory pool.
The memory policy uses this state to build a task-relevant candidate memory view. The online router uses the same state and pre-packing view statistics to select a model; the selected model's budget then determines the final context packing.
After generation, turn-level feedback updates the router, enabling lightweight adaptation without retraining the underlying LLMs.
The QTM prompt, schema, guardrails, and state-to-memory mapping are fixed; the memory view varies by turn, while the router learns online. \method achieves coordination through a shared structured session-state interface.

We evaluate \method on two representative multi-turn benchmarks, \realm~\citep{bian2026realmem} and \mc~\citep{sirdeshmukh2025multichallenge}, using 16 system configurations on each. The experiments test reliability across bounded session interference and long-history memory-driven interaction.
On \realm, \method retains 98.2\% of full-context GPT-4o's persona-macro QA with generation-controller API cost equal to 14.5\% of GPT-4o's generation cost while obtaining the strongest measured memory-recall and retrieval scores.
On full \mc, full-context Qwen2.5 72B is strongest (67/245), followed by GPT-4o (64/245), while \method reaches 58/245. On the two axes most aligned with state recovery and version editing, \method reaches 44/137 versus 45/137 for Qwen2.5 72B at lower recorded cost.
Together, these results support our central claim: coordinating memory exposure and model choice through shared session state provides a useful quality--memory--cost operating point in long-history, memory-dependent interaction.

Our contributions are:
\begin{itemize}[leftmargin=*,topsep=2pt,itemsep=1pt]
    \item We formulate session-level adaptive memory routing as a coordinated sequential decision problem over memory exposure and model selection under a quality--memory--cost trade-off.
    \item We introduce \method, which uses QTM-derived state as a shared control interface for fixed-policy memory construction, online model routing, and feedback updates.
    \item We provide a cross-regime evaluation on two representative multi-turn benchmarks, showing that \method achieves a strong quality--memory--cost trade-off in long-history interaction.
\end{itemize}

\section{Related Work}
\label{sec:related}

\paragraph{LLM routing and online model selection.}
Model routing reduces inference cost by selecting among candidate LLMs instead of using a single expensive model for every request. Prior work studies cascades, benchmark-trained routers, quality-aware routing, preference-trained routers, and dynamic model selection for quality--cost trade-offs~\citep{chen2023frugalgpt,shnitzer2023routing,ding2024hybrid,hu2024routerbench,ong2024routellm,nguyen2024metallm,wang2025mixllm}. Recent work builds on contextual-bandit foundations~\citep{li2010contextual,chu2011contextual,abbasi-yadkori2011} and frames multi-LLM choice as a sequential decision problem with feedback, preferences, or budget constraints~\citep{poon2025online,panda2025adaptive,wei2025learningroutellmsbandit}. These methods primarily route from the current query, prompt, or learned query--model affinity. \method differs in the decision interface: the router operates over structured session state, and this same state also controls memory construction.

\paragraph{Long-term memory, adaptive retrieval, and session evaluation.}
Long-term memory systems help agents store, update, and retrieve information across interactions, from retrieval-augmented generation~\citep{lewis2020retrieval} and generative agents~\citep{park2023generative} to memory-management systems such as MemGPT, Mem0, and A-MEM~\citep{packer2023memgpt,chhikara2025mem0,xu2025amem}. Long-context studies show that simply increasing context length does not guarantee reliable evidence use: agents still struggle with very long conversational history~\citep{maharana-etal-2024-evaluating}, and models can fail depending on where evidence appears in the context~\citep{liu2024lost,bai2023longbench}. Closest to adaptive retrieval, MBA-RAG uses a multi-armed bandit to select retrieval strategies based on question complexity and retrieval cost~\citep{tang2024mbarag}; however, it does not jointly route the generator model or reason over evolving session state. Benchmarks such as \realm and \mc evaluate complementary aspects of memory-driven and path-dependent multi-turn interaction~\citep{bian2026realmem,sirdeshmukh2025multichallenge}. We use them to ask a different question: whether memory exposure and model routing remain reliable when the operating regime shifts from bounded session interference to longer histories, stronger memory pressure, and more difficult retrieval.

\section{Method: Session-Level Adaptive Memory Routing}
\label{sec:method}

We study \emph{session-level adaptive memory routing}: coordinating what memory to expose and which model to invoke during long-term assistant interaction.
At turn $t$, the system observes the current query $q_t$, previous interaction history $H_{<t}$, and a persistent memory store $B_{t-1}$ containing active-task records, recent-turn summaries, and longer-term memories from earlier sessions.
Let $\mathcal{A}$ be the candidate model pool.
Before generating a response, the system builds a candidate memory view $\tilde c_t$, chooses a model $a_t \in \mathcal{A}$, and packs the final context $c_t(a_t)$ for that model.

This formulation makes explicit how simpler systems restrict the decision space.
A fixed configuration uses both a fixed memory policy $\tilde c_t=\pi_{\mathrm{mem}}^{0}(q_t,H_{<t},B_{t-1})$ and a fixed generator $a^0$.
A routing-only system chooses $a_t$ under a memory policy that does not share the router's session-state interface, while a memory-only system adapts $\tilde c_t$ under a fixed generator.
These abstractions are incomplete for session-level interaction because memory construction and $a_t$ interact: the constructed context changes the input difficulty and evidence structure, while the selected model determines how reliably that context is interpreted.

\subsection{Session-State Interface}
\label{sec:state-interface}

\method introduces an explicit state estimation step.
A Query and Topic Manager (\qtm) maps the current turn into a structured session state:
\begin{equation}
    z_t = f_{\qtm}(q_t, H_{<t}, B_{t-1}).
\end{equation}
Rather than producing a free-form summary, QTM emits a constrained control state designed around the two downstream decisions: constructing the candidate memory view $\tilde c_t$ and selecting the model $a_t$.

We organize $z_t$ into three functional groups.
First, session-continuity fields identify whether the turn continues the active thread, shifts topic, or enters a revision, verification, or synthesis phase.
Second, memory-control fields indicate whether prior state is needed, which memory scope should be considered, whether recent history should be preferred, and how much evidence should be exposed.
Third, routing-difficulty fields describe the expected burden placed on the generator, such as memory dependence, ambiguity, and interaction phase; the router later combines these state signals with pre-packing load statistics from the candidate memory view.
These fields are control signals rather than topic labels: the memory policy uses them to determine what parts of $B_{t-1}$ enter $\tilde c_t$, while the router uses them, together with the candidate-view statistics, to estimate which model capacity is appropriate.

The full QTM schema and allowed values are provided in Appendix~\ref{app:qtm_schema}, and the QTM prompt template is provided in Appendix~\ref{app:qtm_prompt}.
The corresponding memory-view and router-feature mappings are detailed in Appendices~\ref{app:prompt_assembly} and~\ref{app:router_features}.
The key design is to use $z_t$ as the shared interface between memory construction and model routing, rather than treating retrieval as a preprocessing step and routing as a separate decision over the raw query.
The QTM prompt, schema, guardrails, and mapping $\psi_{\mathrm{mem}}$ are fixed before evaluation, while the routing policy updates from online feedback. The two decisions are coordinated through a shared structured session-state interface.

\subsection{Coupled Memory Construction and Model Routing}
\label{sec:coupled-decisions}

Given the session state, \method first derives memory-view controls:
\begin{equation}
    m_t = \psi_{\mathrm{mem}}(z_t),
\end{equation}
where $m_t$ summarizes the QTM-induced memory lens, retrieval scope, exposure budget, and recency preference.
The memory policy first constructs a candidate memory view, before model-specific packing:
\begin{equation}
    \tilde c_t = \pi_{\mathrm{mem}}(m_t, q_t, H_{<t}, B_{t-1}).
\end{equation}
In our implementation, the candidate view is assembled from three ordered memory blocks:
\begin{equation}
    \tilde c_t =
    \operatorname{Select}_{m_t}
    \left(
    s_t^{\mathrm{active}},
    h_t^{\mathrm{recent}},
    r_t^{\mathrm{hist}}
    \right),
\end{equation}
where $s_t^{\mathrm{active}}$ stores the active thread or artifact state, $h_t^{\mathrm{recent}}$ contains recent turns, and $r_t^{\mathrm{hist}}$ contains selected historical evidence.
The operator $\operatorname{Select}_{m_t}$ denotes candidate block selection and ordering under the controls in $m_t$, before model-specific prompt-budget packing.
For example, QTM-derived memory need and lens fields decide whether memory is included in the candidate view, retrieval scope selects local active-thread evidence or broader historical memory, top-$k$ hints set the retrieval budget, recency preference biases ranking, and action/phase fields preserve active state for continuation, revision, or verification turns.
Thus, \method separates \emph{access} to memory from \emph{exposure} of memory: a system may store a large history, but the generator receives the selected evidence after packing for its input budget.
Appendix~\ref{app:prompt_assembly} gives the concrete memory-view policy and prompt assembly.

The router uses the candidate view to select a generator; final packing follows that selection:
\begin{align}
    x_t &= \phi(z_t, \tilde c_t, q_t), \\
    a_t &= \pi_t(x_t), \quad a_t \in \mathcal{A}, \\
    c_t(a_t) &= \operatorname{Pack}_{a_t}(\tilde c_t,q_t), \\
    y_t &= g_{a_t}(q_t, c_t(a_t)).
\end{align}
Here $x_t$ is the routing feature vector, $\pi_t$ is the online routing policy, and $\operatorname{Pack}_{a_t}$ clips and assembles the context under the selected model's input budget. The router features describe the pre-packing candidate view, including estimated memory-token load, similarity statistics, and QTM fields; they do not describe the final window-clipped prompt. If generation falls back to another model, the candidate view is repacked for that model. Appendix~\ref{app:prompt_assembly} details budget estimation and prompt assembly.
The router does not memorize raw topic identifiers; instead, it learns state-level patterns such as when memory-heavy revision, temporal recall, or final synthesis requires a stronger operating point.
Appendix~\ref{app:router_features} gives the full feature mapping.

We instantiate $\pi_t$ with a LinUCB-style contextual bandit~\citep{abbasi-yadkori2011}. For each model, $A_a=\gamma I$, $b_a=0$, and $\hat\theta_a=A_a^{-1}b_a$ initially; selection is
\begin{equation}
    a_t = \arg\max_{a \in \mathcal{A}}
    \left(
    \hat{\theta}_a^\top x_t
    + \alpha \sqrt{x_t^\top A_a^{-1} x_t}
    \right),
\end{equation}
where $A_a$ tracks uncertainty, $\gamma>0$ is the ridge parameter, and $\alpha$ controls exploration.
LinUCB is a design choice rather than the central contribution.
The central contribution is the decision interface: the router learns over the same structured state that controls memory exposure.

\subsection{Turn-Level Feedback}
\label{sec:turn-feedback}

After the selected model produces $y_t$, \method estimates a local feedback signal for the routing decision.
Following cost-aware LLM routing and online multi-LLM selection~\citep{chen2023frugalgpt,poon2025online}, we convert turn-level quality into a cost-aware scalar utility:
\begin{equation}
    \hat{r}_t =
    J_{\mathrm{turn}}(y_t, q_t, z_t, c_t(a_t))
    - \lambda C_t^{\mathrm{gen}}(a_t,c_t(a_t)),
    \label{eq:turn_reward}
\end{equation}
where $C_t^{\mathrm{gen}}(a_t,c_t(a_t))$ denotes the selected generator's inference cost, including model-specific pricing and token usage of the final model-specific prompt, and $\lambda$ controls the quality--cost trade-off.

The quality term $J_{\mathrm{turn}}$ is computed by a rubric-based turn judge rather than by a single unstructured score.
The rubric checks whether the response answers the current query, uses the exposed memory consistently, and avoids hard failures such as topic drift or missing required state.
These diagnostics are mapped to a bounded quality score by a fixed weighting and penalty rule, before subtracting generation cost; the rubric, prompt structure, and reward form are provided in Appendix~\ref{app:turn_judge}.

Cost affects model choice through the observed reward rather than through direct price subtraction at decision time.
Expensive model calls are penalized after generation, so future routing decisions gradually favor models that provide better quality--cost utility under similar session states.
Only the selected arm is updated:
\begin{equation}
A_{a_t}\!\leftarrow\!A_{a_t}+x_tx_t^\top,\quad
b_{a_t}\!\leftarrow\!b_{a_t}+\hat r_tx_t,
\label{eq:online_update}
\end{equation}
then $\hat\theta_{a_t}=A_{a_t}^{-1}b_{a_t}$ is recomputed.
The reward is computed only from information available at turn $t$ and does not use future turns, final labels, or benchmark-level pass decisions.

\subsection{Turn-Level Workflow}
\label{sec:turn-workflow}

Algorithm~\ref{alg:mars-workflow} summarizes the turn-level workflow of \method.
We write $\mathcal{I}_t=(q_t,H_{<t},B_{t-1})$ for the turn input.
At each turn, QTM first estimates a structured session state.
The memory policy converts this state into memory-view controls and constructs the candidate view.
The router selects a generator using the shared state and pre-packing candidate statistics; the view is then packed for that model before generation. Turn-level feedback updates the routing policy.
The underlying LLMs are not retrained.

\begin{algorithm}[t]
\small
\caption{\method turn-level workflow}
\label{alg:mars-workflow}
\begin{algorithmic}[1]
\Require turn input $\mathcal{I}_t=(q_t,H_{<t},B_{t-1})$, model pool $\mathcal{A}$, router policy $\pi_t$
\State \textsc{State:} $z_t \leftarrow f_{\qtm}(\mathcal{I}_t)$
\State \textsc{Memory controls:} $m_t \leftarrow \psi_{\mathrm{mem}}(z_t)$
\State \textsc{Candidate view:} $\tilde c_t \leftarrow \pi_{\mathrm{mem}}(m_t,\mathcal{I}_t)$
\State \textsc{Features:} $x_t \leftarrow \phi(z_t,\tilde c_t,q_t)$
\State \textsc{Route:} $a_t \leftarrow \pi_t(x_t)$
\State \textsc{Pack:} $c_t(a_t) \leftarrow \operatorname{Pack}_{a_t}(\tilde c_t,q_t)$
\State \textsc{Generate:} $y_t \leftarrow g_{a_t}(q_t,c_t(a_t))$
\State \textsc{Utility:} $\hat u_t \leftarrow J_{\mathrm{turn}}(y_t,q_t,z_t,c_t(a_t))$
\State \textsc{Reward:} $\hat r_t \leftarrow \hat u_t-\lambda C_t^{\mathrm{gen}}(a_t,c_t(a_t))$
\State \textsc{Router update:} $\pi_{t+1} \leftarrow \operatorname{Update}(\pi_t,x_t,a_t,\hat r_t)$
\State \textsc{Memory update:}
\Statex \hspace{\algorithmicindent} $(H_{\le t},B_t) \leftarrow \operatorname{MemoryUpdate}(\mathcal{I}_t,y_t,z_t)$
\end{algorithmic}
\end{algorithm}

\section{Experimental Setup}
\label{sec:experiments}

Our controlled experiments evaluate \method across two complementary regimes. \realm tests long-history memory-driven interaction; \mc tests bounded structured sessions with interference and path-dependent evaluation. The same 16-system matrix is used on both, enabling consistent comparison across the two regimes.

\subsection{Benchmarks}
\label{sec:benchmarks}

\paragraph{\realm.}
\realm is an external benchmark for real-world memory-driven interaction~\citep{bian2026realmem}. We evaluate its 10-persona evaluation slice: 10 personas and 1,415 queries. We report normalized QA, Mem Recall, Recall@20 (R@20), normalized discounted cumulative gain at 20 (NDCG@20), and API cost. Main quality and memory metrics are macro-averaged over personas; query-weighted aggregates and scored coverage are reported in the appendix.

\paragraph{\mc.}
\mc evaluates state maintenance in bounded but structured multi-turn conversations~\citep{sirdeshmukh2025multichallenge}. Our converted split contains 245 scenarios and 1,480 turns. Each scenario contains intermediate turns followed by a verification turn that uses the official binary rubric question as the final user query. The expected answer is \texttt{YES} when the supplied pass criterion is satisfied, testing whether the system preserved the required topic state and constraints (Appendix~\ref{app:mc_protocol}).

The four axes are \textsc{Inference Memory} (102 scenarios), \textsc{Instruction Retention} (62), \textsc{Self Coherence} (46), and \textsc{Reliable Version Editing} (35). We report the full aggregate first and analyze the first and fourth axes (137 scenarios) separately because they directly test state recovery and local revision. Chunk size 3 bundles trajectories into context blocks with distractor sessions (Appendix~\ref{app:mc_protocol}). For the \method run, we use $\lambda_{\mathrm{cost}}=0.002$, matching the main \realm routing objective.

\subsection{Baselines}
\label{sec:systems}
\label{sec:implementation-feedback}

For both benchmarks, we evaluate 16 systems formed from memory settings and model/router choices. Full-context baselines use one fixed generator for all turns:
GPT-4o~\citep{openai2024gpt4o},
Qwen2.5 72B~\citep{yang2024qwen25},
Mixtral 8$\times$7B Instruct~\citep{jiang2024mixtral},
or Llama3B~\citep{meta2024llama32}. Memory baselines pair \textsc{Mem0}~\citep{chhikara2025mem0} or \textsc{A-MEM}~\citep{xu2025amem} with each fixed generator. Routing baselines include a \textsc{MetaLLM}-style router~\citep{nguyen2024metallm} and hybrids that combine \textsc{Mem0} or \textsc{A-MEM} with \textsc{MetaLLM}-style routing. \method is the full system, using QTM-guided memory construction and online model routing. We use \textsc{Mem0} and \textsc{A-MEM} as representative plug-in memory substrates because they can be consistently crossed with fixed generators and routers across both benchmarks.

QTM uses Qwen2.5-7B-Instruct~\citep{yang2024qwen25}
with a fixed prompt, schema, guardrails, and state-to-memory mapping.
A GPT-4o-mini~\citep{openai2024gpt4omini} turn judge supplies
local reward after generation; it never sees future turns,
final labels, gold final answers, or benchmark pass decisions
(Appendix~\ref{app:turn_judge}).

\subsection{Ablation Protocol}
\label{sec:ablation-protocol}

We run diagnostic ablations on a fixed three-persona \realm evaluation slice containing 473 turns. All variants use the same candidate model pool, deterministic generation, and a matched top-10 historical retrieval budget. The slice uses $\lambda_{\mathrm{cost}}=0.5$, compared with $\lambda_{\mathrm{cost}}=0.002$ in the main \realm run; its retrieval budget also differs from the main policy's top-$k$ hints. Comparisons are therefore made within this diagnostic configuration. The ablations examine QTM-guided context construction versus raw-query semantic retrieval, QTM state/header exposure in the generator prompt, QTM-derived routing versus query-embedding routing with semantic or recency-based memory, and online adaptation via a no-update diagnostic. The no-update variant freezes the router at initialization, so we treat it as a stress test of online adaptation rather than as a tuned static-router baseline. Ablation metrics follow the official \realm evaluation pipeline, with QA normalized to the same scale as the main results. Full settings and diagnostics are provided in Appendix~\ref{app:ablations}.

\subsection{Evaluation, Cost, and Reporting}
\label{sec:evaluation-protocols}
\label{sec:cost-reporting}

For \realm, we use the official retrieval and generation evaluation pipeline, with GPT-4o as the final judge when judging is required. This final evaluator is separate from the GPT-4o-mini turn-level judge used for online routing updates. For \mc, we use a GPT-4o path-aware final judge following the benchmark style: it checks trajectory continuity and final-answer correctness. A scenario passes only when both checks are accepted and the model's parseable final YES/NO answer matches the expected criterion (Appendix~\ref{app:final_judges}).

For \realm, main tables report generator and QTM/controller API cost, excluding online feedback; Results separately reports reconstructed feedback-inclusive totals. For \mc, the archived runs did not itemize QTM/controller cost separately, so tables retain the original run-level recorded-cost convention. Final benchmark evaluators and local memory compute and storage are excluded from these API cost views. The online reward uses only selected-generator cost (Appendix~\ref{app:cost_accounting}). We bootstrap \realm by persona and \mc by scenario, and separately test eight chunk orders on 30 fixed \mc scenarios; Appendix~\ref{app:uncertainty_order} gives these diagnostics.

\section{Results}
\label{sec:results}

We evaluate the quality--memory--cost trade-off across long-history interaction on \realm and bounded session interference on \mc.

\subsection{\realm: QA--Cost Trade-off and Memory Metrics}
\label{sec:results-realmem}

\begin{figure}[htbp]
\centering
\includegraphics[width=\linewidth]{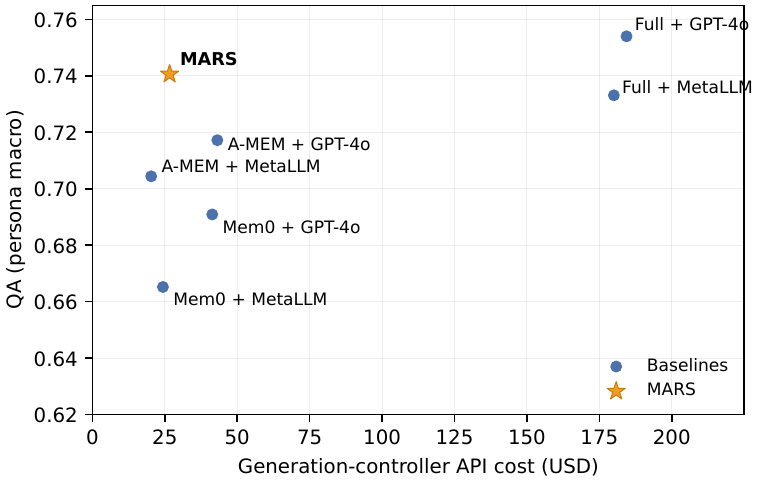}
\caption{\realm persona-macro QA versus generation-controller API cost for key systems. Online feedback and final evaluation are excluded.}
\label{fig:realmem-qacost}
\end{figure}

Table~\ref{tab:realmem-main} and Figure~\ref{fig:realmem-qacost} report persona-macro quality and generation-controller API cost over 10 personas. Appendix~\ref{app:realmem-full} gives all 16 systems.

\begin{table}[htbp]
\centering
\small
\setlength{\tabcolsep}{7pt}
\renewcommand{\arraystretch}{1.08}
\begin{adjustbox}{max width=\linewidth}
\begin{tabular}{@{}llccccc@{}}
\toprule
Memory Setting & Model / Router 
& QA $\uparrow$ 
& Mem Rec. $\uparrow$ 
& R@20 $\uparrow$ 
& NDCG@20 $\uparrow$ 
& Gen.-Ctrl. (\$) $\downarrow$ \\
\midrule
\multicolumn{2}{c}{\textbf{\method}}
& \underline{0.7406} & \textbf{0.6794} & \textbf{0.8015} & \textbf{0.6168} & 26.6480 \\
\midrule
Full context & GPT-4o 
& \textbf{0.7540} 
& 0.1640 & 0.5188 & 0.3216 & 184.3967 \\
Full context & \textsc{MetaLLM} 
& 0.7331 & 0.1603 & 0.5188 & 0.3216 & 180.0189 \\
\textsc{Mem0} & GPT-4o 
& 0.6909 & 0.5997 & 0.6305 & 0.5388 & 41.3834 \\
\textsc{Mem0} & \textsc{MetaLLM} 
& 0.6652 & 0.6048 & 0.6305 & 0.5388 & 24.3499 \\
\textsc{A-MEM} & GPT-4o 
& 0.7172 & \underline{0.6479} & \underline{0.6817} & \underline{0.5846} & 43.1311 \\
\textsc{A-MEM} & \textsc{MetaLLM} 
& 0.7044 & 0.6471 & 0.6817 & 0.5846 & 20.2875 \\
\bottomrule
\end{tabular}
\end{adjustbox}
\caption{\realm key systems (10 personas; 1,415-query pool). Metrics are persona-macro averages. Cost records generation and controller API calls; online feedback and final evaluation are excluded. Overflow-dominated full-context rows remain in Appendix~\ref{app:realmem-full}.}
\label{tab:realmem-main}
\end{table}

Full-context GPT-4o obtains 0.7540 QA at \$184.40. \method reaches 0.7406 (98.2\%) at \$26.65 (14.5\%) under the generation-controller view and has the best memory-recall and retrieval scores. A-MEM+MetaLLM is cheaper (\$20.29) but has lower QA (0.7044) and memory metrics, giving a different operating point. Adding the estimated GPT-4o-mini feedback cost gives \method a feedback-inclusive API total of approximately \$32.53 (17.6\% of full-context GPT-4o), compared with approximately \$23.49 for A-MEM+MetaLLM under the same accounting. These estimates use logged input-token counts and estimated output usage; final benchmark evaluation and local compute are excluded (Appendix~\ref{app:cost_accounting}).

The persona-bootstrap 95\% interval for QA is [0.7183, 0.7641]; other intervals are in Appendix~\ref{app:uncertainty_order}. Archived provider errors show context-limit failures on 1,344/1,415 queries for both Qwen2.5-72B and Mixtral, and 1,247/1,415 for Llama3B; affected rows receive zero score and zero recorded cost. Their low QA therefore reflects the evaluated full-history configuration's context limits as well as response quality.

\subsection{\mc: Full Aggregate and Mechanism-Focused Subset}
\label{sec:results-mc}

Table~\ref{tab:mc-main} reports the full aggregate first and the mechanism-focused subset second; complete matrices are in Appendices~\ref{app:mc-full}--\ref{app:mc-target}.

\begin{table}[htbp]
\centering
\footnotesize
\setlength{\tabcolsep}{3pt}
\begin{adjustbox}{max width=\linewidth}
\begin{tabular}{@{}lccc@{}}
\toprule
System & Full pass $\uparrow$ & Subset pass $\uparrow$ & \makecell{Recorded subset\\cost (\$) $\downarrow$}\\
\midrule
\textbf{\method} & 58/245 & \underline{44/137} & 1.5670 \\
Full Qwen2.5 72B & \textbf{67/245} & \textbf{45/137} & 1.8528\\
Full GPT-4o & \underline{64/245} & 36/137 & 9.1044\\
\textsc{Mem0}+Mixtral & 57/245 & 39/137 & 0.4620\\
\bottomrule
\end{tabular}
\end{adjustbox}
\caption{\mc full aggregate and mechanism-focused subset. Costs follow the original run-level accounting convention; QTM/controller cost was not separately itemized in the archived logs.}
\label{tab:mc-main}
\end{table}

On full \mc, Qwen2.5 72B (67/245) and GPT-4o (64/245) outperform \method (58/245). Qwen's full-set recorded cost is also slightly lower (\$3.1665 versus \$3.2504). On the focused subset, \method reaches 44/137 versus 45/137 for Qwen at lower recorded cost. Per-axis, \method remains close on \textsc{Inference Memory} (33/102 versus 36/102) and is higher on \textsc{Reliable Version Editing} (11/35 versus 9/35), while larger gaps remain on the other two axes (Appendix~\ref{app:mc-axis-breakdown}). This supports using the subset as a mechanism-focused view of state recovery and local revision rather than a replacement for the full aggregate. \method's full-set pass rate is 0.2367 with a scenario-bootstrap 95\% interval of [0.1837, 0.2898]. Across eight randomized chunk orders of the same fixed 30-scenario subset, with a freshly initialized bandit for each replay, the diagnostic pass rate averages 18.3\% with a sample standard deviation of 3.6 percentage points (4/30--7/30); Appendix~\ref{app:uncertainty_order} gives the construction and per-order results.

\subsection{\realm Diagnostic Ablations}
\label{sec:results-ablation}

We run diagnostic ablations on a fixed three-persona \realm evaluation slice (473 turns), with matched top-10 retrieval and the same model pool.

\begin{table}[htbp]
\centering
\small
\setlength{\tabcolsep}{4pt}
\renewcommand{\arraystretch}{1.05}
\begin{adjustbox}{max width=\linewidth}
\begin{tabular}{lrrrr}
\toprule
Variant & QA $\uparrow$ & Mem Rec. $\uparrow$ & NDCG@10 $\uparrow$ & \makecell{Gen.-Ctrl.\\(\$) $\downarrow$} \\
\midrule
Full \method & \textbf{0.7444} & \textbf{0.6804} & \textbf{0.5974} & 5.643 \\
Embed. semantic & 0.7231 & 0.6429 & 0.5162 & 5.911 \\
No state/header & 0.7099 & 0.6397 & 0.5162 & 6.338 \\
Recent-only & 0.6279 & 0.1070 & 0.2299 & 5.638 \\
No-QTM semantic & \underline{0.7387} & 0.6287 & 0.5162 & \textbf{4.570} \\
No-QTM recent & 0.6244 & 0.1118 & 0.2299 & 4.918 \\
No update$^\dagger$ & 0.4221 & 0.6882 & 0.5983 & 0.439 \\
\bottomrule
\end{tabular}
\end{adjustbox}
\caption{\realm diagnostic ablations (473 turns; $\lambda_{\mathrm{cost}}=0.5$). Bold marks the best online variant. $^\dagger$No-update freezes the untuned cold-start router. Full metrics: Appendix~\ref{app:ablations}.}
\label{tab:realmem-ablation-main}
\end{table}

Full \method has the highest QA among online variants. Under the matched top-10 budget, embedding-only semantic retrieval lowers QA from 0.7444 to 0.7231 and Mem Recall from 0.6804 to 0.6429, while recent-only memory reduces Mem Recall to 0.1070. Conversely, the no-update row retains strong retrieval metrics but drops QA to 0.4221. These diagnostics support complementary roles for state-guided memory construction and online adaptation.

\subsection{QTM Prompt and Controller Sensitivity}
\label{sec:qtm-prompt-sensitivity}

We probe local sensitivity of the QTM interface under two types of controlled variation on the same 150 frozen session snapshots from seven \realm evaluation personas. With Qwen2.5-7B-Instruct fixed, we compare the original prompt with a semantics-preserving paraphrase and an instruction-reordered variant. We then keep the original prompt and schema fixed and replace the QTM controller with
Mistral-7B-Instruct-v0.3~\citep{mistralai2024mistral7bv03}. Across all comparisons, the session history, memory pool, retrieval policy, and deterministic decoding are held fixed. We measure agreement of the effective \texttt{MATCH}/\texttt{NEW\_TOPIC} decision and Jaccard overlap of the pre-packing retrieved memory-source IDs.

\begin{table}[htbp]
\centering
\small
\setlength{\tabcolsep}{5pt}
\renewcommand{\arraystretch}{1.05}
\begin{adjustbox}{max width=\linewidth}
\begin{tabular}{@{}lcc@{}}
\toprule
Controlled variation & Action agr. $\uparrow$ & Retrieval Jaccard $\uparrow$ \\
\midrule
Prompt paraphrase & 90.0\% & 0.920 \\
Instruction reordering & 96.0\% & 0.973 \\
Mistral-7B controller & 82.7\% & 0.875 \\
\bottomrule
\end{tabular}
\end{adjustbox}
\caption{QTM interface-sensitivity probe over 150 fixed session snapshots. Prompt variants use the default Qwen2.5-7B-Instruct controller; the controller-swap row uses the original prompt/schema with Mistral-7B-Instruct-v0.3. Each row is compared with the Qwen2.5-7B-Instruct original-prompt reference.}
\label{tab:qtm-prompt-sensitivity-main}
\end{table}

The prompt perturbations preserve most effective thread-boundary decisions and retrieved memory sources; on the 60 inputs with active thread history---where a \texttt{NEW\_TOPIC} decision can trigger archival---agreement rises to 95.0\%/98.3\% and retrieval Jaccard to 0.984/1.000 for paraphrasing/reordering, respectively. Under controller substitution, the effective action agrees on 124/150 inputs (82.7\%) with retrieval Jaccard 0.875; on the same 60 active-history inputs, agreement is 53/60 (88.3\%) and retrieval Jaccard is 0.960. Together, these probes provide complementary evidence that the key thread-boundary and retrieval decisions remain largely preserved under the tested prompt changes and one same-scale controller substitution. Appendices~\ref{app:qtm-prompt-sensitivity}--\ref{app:qtm-controller-swap} provide detailed diagnostics.

\subsection{Cross-Regime Analysis}
\label{sec:results-cross}

Static Qwen2.5 72B leads on full \mc, while full-context Qwen2.5-72B and Mixtral exceed their context limits on 1,344/1,415 \realm queries. This contrast suggests a greater role for selective memory exposure under long-history pressure; strong static models remain effective in bounded contexts.

\section{Conclusion}
\label{sec:conclusion}

We introduced \method, which coordinates a fixed state-conditioned memory policy and an online model router through a shared session-state interface. It achieves a strong quality--memory--cost trade-off on \realm and remains competitive on the mechanism-focused \mc subset, while trailing the two strongest static systems on the full aggregate.

\newpage
\bibliographystyle{acl_natbib}
\bibliography{custom}

@inproceedings{maharana-etal-2024-evaluating,
  title = {Evaluating Very Long-Term Conversational Memory of {LLM} Agents},
  author = {Maharana, Adyasha and Lee, Dong-Ho and Tulyakov, Sergey and Bansal, Mohit and Barbieri, Francesco and Fang, Yuwei},
  booktitle = {Proceedings of the 62nd Annual Meeting of the Association for Computational Linguistics (Volume 1: Long Papers)},
  editor = {Ku, Lun-Wei and Martins, Andre and Srikumar, Vivek},
  year = {2024},
  month = aug,
  address = {Bangkok, Thailand},
  publisher = {Association for Computational Linguistics},
  pages = {13851--13870},
  doi = {10.18653/v1/2024.acl-long.747},
  url = {https://aclanthology.org/2024.acl-long.747/}
}

@article{chen2023frugalgpt,
  title = {FrugalGPT: How to Use Large Language Models While Reducing Cost and Improving Performance},
  author = {Chen, Lingjiao and Zaharia, Matei and Zou, James},
  journal = {arXiv preprint arXiv:2305.05176},
  year = {2023},
  doi = {10.48550/arXiv.2305.05176},
  url = {https://arxiv.org/abs/2305.05176}
}

@article{ong2024routellm,
  title = {RouteLLM: Learning to Route LLMs with Preference Data},
  author = {Ong, Isaac and Almahairi, Amjad and Wu, Vincent and Chiang, Wei-Lin and Wu, Tianhao and Gonzalez, Joseph E. and Kadous, M. Waleed and Stoica, Ion},
  journal = {arXiv preprint arXiv:2406.18665},
  year = {2024},
  doi = {10.48550/arXiv.2406.18665},
  url = {https://arxiv.org/abs/2406.18665}
}

@article{nguyen2024metallm,
  title = {MetaLLM: A High-performant and Cost-efficient Dynamic Framework for Wrapping LLMs},
  author = {Nguyen, Quang H. and Hoang, Duy C. and Decugis, Juliette and Manchanda, Saurav and Chawla, Nitesh V. and Doan, Khoa D.},
  journal = {arXiv preprint arXiv:2407.10834},
  year = {2024},
  doi = {10.48550/arXiv.2407.10834},
  url = {https://arxiv.org/abs/2407.10834}
}

@article{wang2025mixllm,
  title = {MixLLM: Dynamic Routing in Mixed Large Language Models},
  author = {Wang, Xinyuan and Liu, Yanchi and Cheng, Wei and Zhao, Xujiang and Chen, Zhengzhang and Yu, Wenchao and Fu, Yanjie and Chen, Haifeng},
  journal = {arXiv preprint arXiv:2502.18482},
  year = {2025},
  doi = {10.48550/arXiv.2502.18482},
  url = {https://arxiv.org/abs/2502.18482}
}

@article{poon2025online,
  title = {Online Multi-LLM Selection via Contextual Bandits under Unstructured Context Evolution},
  author = {Poon, Manhin and Dai, XiangXiang and Liu, Xutong and Kong, Fang and Lui, John C. S. and Zuo, Jinhang},
  journal = {arXiv preprint arXiv:2506.17670},
  year = {2025},
  doi = {10.48550/arXiv.2506.17670},
  url = {https://arxiv.org/abs/2506.17670}
}

@article{chhikara2025mem0,
  title = {Mem0: Building Production-Ready AI Agents with Scalable Long-Term Memory},
  author = {Chhikara, Prateek and Khant, Dev and Aryan, Saket and Singh, Taranjeet and Yadav, Deshraj},
  journal = {arXiv preprint arXiv:2504.19413},
  year = {2025},
  doi = {10.48550/arXiv.2504.19413},
  url = {https://arxiv.org/abs/2504.19413}
}

@article{xu2025amem,
  title = {A-MEM: Agentic Memory for LLM Agents},
  author = {Xu, Wujiang and Liang, Zujie and Mei, Kai and Gao, Hang and Tan, Juntao and Zhang, Yongfeng},
  journal = {arXiv preprint arXiv:2502.12110},
  year = {2025},
  doi = {10.48550/arXiv.2502.12110},
  url = {https://arxiv.org/abs/2502.12110}
}

@article{bian2026realmem,
  title = {RealMem: Benchmarking LLMs in Real-World Memory-Driven Interaction},
  author = {Bian, Haonan and Yao, Zhiyuan and Hu, Sen and Xu, Zishan and Zhang, Shaolei and Guo, Yifu and Yang, Ziliang and Han, Xueran and Wang, Huacan and Chen, Ronghao},
  journal = {arXiv preprint arXiv:2601.06966},
  year = {2026},
  doi = {10.48550/arXiv.2601.06966},
  url = {https://arxiv.org/abs/2601.06966}
}

@inproceedings{abbasi-yadkori2011,
  title = {Improved Algorithms for Linear Stochastic Bandits},
  author = {Abbasi-Yadkori, Yasin and P{\'a}l, D{\'a}vid and Szepesv{\'a}ri, Csaba},
  booktitle = {Advances in Neural Information Processing Systems 24},
  year = {2011},
  url = {https://proceedings.neurips.cc/paper_files/paper/2011/hash/e1d5be1c7f2f456670de3d53c7b54f4a-Abstract.html}
}

@article{sirdeshmukh2025multichallenge,
  title = {MultiChallenge: A Realistic Multi-Turn Conversation Evaluation Benchmark Challenging to Frontier {LLM}s},
  author = {Sirdeshmukh, Ved and Deshpande, Kaustubh and Mols, Johannes and Jin, Lifeng and Cardona, Ed-Yeremai and Lee, Dean and Kritz, Jeremy and Primack, Willow and Yue, Summer and Xing, Chen},
  journal = {arXiv preprint arXiv:2501.17399},
  year = {2025},
  doi = {10.48550/arXiv.2501.17399},
  url = {https://arxiv.org/abs/2501.17399}
}

@article{ding2024hybrid,
  title = {Hybrid {LLM}: Cost-Efficient and Quality-Aware Query Routing},
  author = {Ding, Dujian and Mallick, Ankur and Wang, Chi and Sim, Robert and Mukherjee, Subhabrata and Ruhle, Victor and Lakshmanan, Laks V. S. and Awadallah, Ahmed Hassan},
  journal = {arXiv preprint arXiv:2404.14618},
  year = {2024},
  doi = {10.48550/arXiv.2404.14618},
  url = {https://arxiv.org/abs/2404.14618}
}

@article{hu2024routerbench,
  title = {RouterBench: A Benchmark for Multi-{LLM} Routing System},
  author = {Hu, Qitian Jason and Bieker, Jacob and Li, Xiuyu and Jiang, Nan and Keigwin, Benjamin and Ranganath, Gaurav and Keutzer, Kurt and Upadhyay, Shriyash Kaustubh},
  journal = {arXiv preprint arXiv:2403.12031},
  year = {2024},
  doi = {10.48550/arXiv.2403.12031},
  url = {https://arxiv.org/abs/2403.12031}
}

@inproceedings{lewis2020retrieval,
  title = {Retrieval-Augmented Generation for Knowledge-Intensive {NLP} Tasks},
  author = {Lewis, Patrick and Perez, Ethan and Piktus, Aleksandra and Petroni, Fabio and Karpukhin, Vladimir and Goyal, Naman and K{\"u}ttler, Heinrich and Lewis, Mike and Yih, Wen-tau and Rockt{\"a}schel, Tim and Riedel, Sebastian and Kiela, Douwe},
  booktitle = {Advances in Neural Information Processing Systems},
  volume = {33},
  pages = {9459--9474},
  year = {2020},
  url = {https://proceedings.neurips.cc/paper/2020/hash/6b493230205f780e1bc26945df7481e5-Abstract.html}
}

@inproceedings{park2023generative,
  title = {Generative Agents: Interactive Simulacra of Human Behavior},
  author = {Park, Joon Sung and O'Brien, Joseph C. and Cai, Carrie J. and Morris, Meredith Ringel and Liang, Percy and Bernstein, Michael S.},
  booktitle = {Proceedings of the 36th Annual ACM Symposium on User Interface Software and Technology},
  series = {UIST '23},
  year = {2023},
  publisher = {Association for Computing Machinery},
  doi = {10.1145/3586183.3606763},
  url = {https://doi.org/10.1145/3586183.3606763}
}

@article{packer2023memgpt,
  title = {{MemGPT}: Towards {LLM}s as Operating Systems},
  author = {Packer, Charles and Wooders, Sarah and Lin, Kevin and Fang, Vivian and Patil, Shishir G. and Stoica, Ion and Gonzalez, Joseph E.},
  journal = {arXiv preprint arXiv:2310.08560},
  year = {2023},
  doi = {10.48550/arXiv.2310.08560},
  url = {https://arxiv.org/abs/2310.08560}
}

@article{liu2024lost,
  title = {Lost in the Middle: How Language Models Use Long Contexts},
  author = {Liu, Nelson F. and Lin, Kevin and Hewitt, John and Paranjape, Ashwin and Bevilacqua, Michele and Petroni, Fabio and Liang, Percy},
  journal = {Transactions of the Association for Computational Linguistics},
  volume = {12},
  pages = {157--173},
  year = {2024},
  doi = {10.1162/tacl_a_00638},
  url = {https://doi.org/10.1162/tacl_a_00638}
}

@article{bai2023longbench,
  title = {{LongBench}: A Bilingual, Multitask Benchmark for Long Context Understanding},
  author = {Bai, Yushi and Lv, Xin and Zhang, Jiajie and Lyu, Hongchang and Tang, Jiankai and Huang, Zhidian and Du, Zhengxiao and Liu, Xiao and Zeng, Aohan and Hou, Lei and Dong, Yuxiao and Tang, Jie and Li, Juanzi},
  journal = {arXiv preprint arXiv:2308.14508},
  year = {2023},
  doi = {10.48550/arXiv.2308.14508},
  url = {https://arxiv.org/abs/2308.14508}
}

@article{shnitzer2023routing,
  title = {Large Language Model Routing with Benchmark Datasets},
  author = {Shnitzer, Tal and Ou, Anthony and Silva, M{\'i}rian and Soule, Kate and Sun, Yuekai and Solomon, Justin and Thompson, Neil and Yurochkin, Mikhail},
  journal = {arXiv preprint arXiv:2309.15789},
  year = {2023},
  doi = {10.48550/arXiv.2309.15789},
  url = {https://arxiv.org/abs/2309.15789}
}

@inproceedings{li2010contextual,
  title = {A Contextual-Bandit Approach to Personalized News Article Recommendation},
  author = {Li, Lihong and Chu, Wei and Langford, John and Schapire, Robert E.},
  booktitle = {Proceedings of the 19th International Conference on World Wide Web},
  series = {WWW '10},
  pages = {661--670},
  year = {2010},
  publisher = {Association for Computing Machinery},
  doi = {10.1145/1772690.1772758},
  url = {https://doi.org/10.1145/1772690.1772758}
}

@inproceedings{chu2011contextual,
  title = {Contextual Bandits with Linear Payoff Functions},
  author = {Chu, Wei and Li, Lihong and Reyzin, Lev and Schapire, Robert E.},
  booktitle = {Proceedings of the Fourteenth International Conference on Artificial Intelligence and Statistics},
  series = {Proceedings of Machine Learning Research},
  volume = {15},
  pages = {208--214},
  year = {2011},
  publisher = {PMLR},
  url = {https://proceedings.mlr.press/v15/chu11a.html}
}

@article{panda2025adaptive,
  title = {Adaptive LLM Routing under Budget Constraints},
  author = {Panda, Pranoy and Magazine, Raghav and Devaguptapu, Chaitanya and Takemori, Sho and Sharma, Vishal},
  journal = {arXiv preprint arXiv:2508.21141},
  year = {2025},
  doi = {10.48550/arXiv.2508.21141},
  url = {https://arxiv.org/abs/2508.21141}
}

@misc{wei2025learningroutellmsbandit,
      title={Learning to Route {LLM}s from Bandit Feedback: One Policy, Many Trade-offs}, 
      author={Wang Wei and Tiankai Yang and Hongjie Chen and Yue Zhao and Franck Dernoncourt and Ryan A. Rossi and Hoda Eldardiry},
      year={2025},
      eprint={2510.07429},
      archivePrefix={arXiv},
      primaryClass={cs.LG},
      url={https://arxiv.org/abs/2510.07429}, 
}

@article{tang2024mbarag,
  title = {{MBA-RAG}: a Bandit Approach for Adaptive Retrieval-Augmented Generation through Question Complexity},
  author = {Tang, Xiaqiang and Gao, Qiang and Li, Jian and Du, Nan and Li, Qi and Xie, Sihong},
  journal = {arXiv preprint arXiv:2412.01572},
  year = {2024},
  doi = {10.48550/arXiv.2412.01572},
  url = {https://arxiv.org/abs/2412.01572}
}

@misc{openai2024gpt4o,
  author = {{OpenAI}},
  title = {Hello {GPT-4o}},
  year = {2024},
  month = may,
  url = {https://openai.com/index/hello-gpt-4o/},
  note = {Accessed October 9, 2026}
}

@misc{openai2024gpt4omini,
  author = {{OpenAI}},
  title = {{GPT-4o mini}: advancing cost-efficient intelligence},
  year = {2024},
  month = jul,
  url = {https://openai.com/index/gpt-4o-mini-advancing-cost-efficient-intelligence/},
  note = {Accessed October 9, 2026}
}

@article{yang2024qwen25,
  author = {An Yang and Baosong Yang and Beichen Zhang and
    Binyuan Hui and Bo Zheng and Bowen Yu and Chengyuan Li and
    Dayiheng Liu and Fei Huang and Haoran Wei and Huan Lin and
    Jian Yang and Jianhong Tu and Jianwei Zhang and Jianxin Yang and
    Jiaxi Yang and Jingren Zhou and Junyang Lin and Kai Dang and
    Keming Lu and Keqin Bao and Kexin Yang and Le Yu and Mei Li and
    Mingfeng Xue and Pei Zhang and Qin Zhu and Rui Men and Runji Lin and
    Tianhao Li and Tingyu Xia and Xingzhang Ren and Xuancheng Ren and
    Yang Fan and Yang Su and Yichang Zhang and Yu Wan and Yuqiong Liu and
    Zeyu Cui and Zhenru Zhang and Zihan Qiu},
  title = {{Qwen2.5} Technical Report},
  journal = {arXiv preprint arXiv:2412.15115},
  year = {2024},
  url = {https://arxiv.org/abs/2412.15115v1}
}

@article{jiang2024mixtral,
  author = {Albert Q. Jiang and Alexandre Sablayrolles and
    Antoine Roux and Arthur Mensch and Blanche Savary and
    Chris Bamford and Devendra Singh Chaplot and Diego de las Casas and
    Emma Bou Hanna and Florian Bressand and Gianna Lengyel and
    Guillaume Bour and Guillaume Lample and L{\'e}lio Renard Lavaud and
    Lucile Saulnier and Marie-Anne Lachaux and Pierre Stock and
    Sandeep Subramanian and Sophia Yang and Szymon Antoniak and
    Teven Le Scao and Th{\'e}ophile Gervet and Thibaut Lavril and
    Thomas Wang and Timoth{\'e}e Lacroix and William El Sayed},
  title = {{Mixtral} of Experts},
  journal = {arXiv preprint arXiv:2401.04088},
  year = {2024},
  doi = {10.48550/arXiv.2401.04088},
  url = {https://arxiv.org/abs/2401.04088}
}

@misc{meta2024llama32,
  author = {{Meta}},
  title = {{Llama 3.2} Model Card},
  year = {2024},
  url = {https://github.com/meta-llama/llama-models/blob/main/models/llama3_2/MODEL_CARD.md},
  note = {Accessed October 9, 2026}
}

@misc{mistralai2024mistral7bv03,
  author = {{Mistral AI}},
  title = {Model Card for {Mistral-7B-Instruct-v0.3}},
  year = {2024},
  month = may,
  url = {https://huggingface.co/mistralai/Mistral-7B-Instruct-v0.3},
  note = {Accessed October 9, 2026}
}

@inproceedings{xiao2024cpack,
  author = {Xiao, Shitao and Liu, Zheng and Zhang, Peitian and
    Muennighoff, Niklas and Lian, Defu and Nie, Jian-Yun},
  title = {{C-Pack}: Packed Resources For General {Chinese} Embeddings},
  booktitle = {Proceedings of the 47th International ACM SIGIR
    Conference on Research and Development in Information Retrieval},
  year = {2024},
  pages = {641--649},
  publisher = {Association for Computing Machinery},
  doi = {10.1145/3626772.3657878},
  url = {https://doi.org/10.1145/3626772.3657878}
}

\clearpage
\appendix
\section{Appendix Overview}
\label{app:overview}

This appendix provides implementation details, benchmark construction, full result tables, and diagnostic analyses for \method. Appendix~\ref{app:implementation} specifies the \qtm state interface, memory-view construction, router features, turn-level reward, and cost accounting. Appendix~\ref{app:benchmark_details} describes the \realm and \mc protocols, including metric normalization and the chunked-context construction used for \mc. Appendix~\ref{app:additional_results} reports full system matrices. Appendix~\ref{app:ablations} reports the three-persona \realm component ablations and cost-weight sensitivity, together with QTM prompt- and controller-sensitivity probes. Appendix~\ref{app:qualitative} provides anonymized qualitative examples and routing-behavior diagnostics from fixed evaluation outputs. Appendix~\ref{app:deployment_repro} discusses deployment feedback and reproducibility information.

\section{Implementation Details}
\label{app:implementation}

The main text formulates \method as coordinating memory exposure and model routing through shared state. Its QTM prompt, schema, guardrails, and state-to-memory mapping are fixed; only the router updates online.

\paragraph{Optimization scope.} In the evaluated implementation, the QTM schema, guardrails, and state-to-memory policy are fixed before evaluation, while the LinUCB router adapts online. Coordination is realized through the shared session state and the turn-dependent memory view induced by that state.

\FloatBarrier
\subsection{QTM Schema and State Representation}
\label{app:qtm_schema}

The Query and Topic Manager (\qtm) maps the current turn into a structured session state. In our implementation, \qtm is instantiated with \texttt{qwen/qwen-2.5-7b-instruct}. We use this smaller instruction model because \qtm does not generate final answers; it produces a compact control state used by both the memory policy and the online router.

The schema describes task state rather than topic identity. Omitting raw topic labels directs the router toward state-level patterns, such as the model utility associated with memory-heavy revision, temporal recall, or final synthesis. Table~\ref{tab:qtm-schema} summarizes the fields used by the current implementation.

\begin{table}[htbp]
\centering
\small
\setlength{\tabcolsep}{5pt}
\renewcommand{\arraystretch}{1.12}
\begin{tabularx}{\textwidth}{@{}P{0.25\textwidth}YccY@{}}
\toprule
Field & Allowed values / representation & Memory & Router & Role in the decision interface \\
\midrule
\texttt{action} & \texttt{MATCH}, \texttt{NEW\_\allowbreak TOPIC} & yes & yes & Segments the turn as continuation of the active thread or a topic switch. \\
\texttt{query\_\allowbreak type} & \texttt{STATIC}, \texttt{DYNAMIC}, \texttt{TEMPORAL}, \texttt{PROACTIVE} & yes & yes & Distinguishes self-contained, evolving, temporal, and proactive turns. \\
\texttt{phase} & \texttt{EXPLORE}, \texttt{CONSTRAIN}, \texttt{REVISE}, \texttt{SYNTHESIZE}, \texttt{VERIFY} & yes & yes & Encodes interaction phase for memory exposure and model selection. \\
\texttt{taxonomy.*} & \texttt{need\_\allowbreak memory}, \texttt{memory\_\allowbreak span}, ambiguity, and risk metadata & yes & yes & Records decision-relevant memory and ambiguity state without raw topic identity. \\
\texttt{retrieval\_\allowbreak scope} & \texttt{LOCAL}, \texttt{GLOBAL} & yes & yes & Selects active-thread evidence or broader historical retrieval. \\
\texttt{top\_\allowbreak k\_\allowbreak hint} & \texttt{3}, \texttt{5}, \texttt{8} & yes & yes & Gives the memory constructor a retrieval-budget hint. \\
\texttt{prefer\_\allowbreak recent} & \texttt{0}, \texttt{1} & yes & yes & Biases memory construction toward recent history when relevant. \\
\texttt{lens\_\allowbreak recommendation} & \texttt{0} or \texttt{3} & yes & yes & Controls minimal or full memory exposure. \\
\texttt{lens\_\allowbreak confidence}, \texttt{lens\_\allowbreak reason} & confidence and short reason & yes & optional & Provides a traceable rationale for the selected memory lens. \\
\texttt{qtm\_\allowbreak llm\_\allowbreak called}, \texttt{qtm\_\allowbreak usage} & logged diagnostics & no & no & Used for cost tracing and QTM-overhead accounting. \\
\bottomrule
\end{tabularx}
\caption{QTM state schema used in the \realm implementation. The main-text abstract fields map to this concrete schema: for example, memory need is represented primarily through \texttt{taxonomy.need\_\allowbreak memory}; memory view is represented through \texttt{lens\_\allowbreak recommendation}, \texttt{retrieval\_\allowbreak scope}, and \texttt{top\_\allowbreak k\_\allowbreak hint}.}
\label{tab:qtm-schema}
\end{table}

\begin{figure}[htbp]
\centering
\fbox{%
\begin{minipage}{0.96\linewidth}
\small
\textbf{QTM structured output shape used in \realm mode.}

\smallskip
\noindent\textbf{Inputs:} \texttt{ACTIVE CONTEXT}, \texttt{USER INPUT}, and \texttt{SYSTEM RULES}. The QTM emits controller state only and does not generate the final answer.

\vspace{0.6em}
\noindent\textbf{Required JSON fields:}\par\smallskip
\renewcommand{\arraystretch}{1.12}
\noindent\begin{tabularx}{\linewidth}{@{}P{0.31\linewidth}X@{}}
\texttt{action}: & \texttt{MATCH} or \texttt{NEW\_\allowbreak TOPIC} \\
\texttt{query\_\allowbreak type}: & \texttt{STATIC}, \texttt{DYNAMIC}, \texttt{TEMPORAL}, or \texttt{PROACTIVE} \\
\texttt{phase}: & \texttt{EXPLORE}, \texttt{CONSTRAIN}, \texttt{REVISE}, \texttt{SYNTHESIZE}, or \texttt{VERIFY} \\
\texttt{taxonomy}:
& memory and risk metadata, including \texttt{need\_\allowbreak memory}, \texttt{memory\_\allowbreak span}, and ambiguity \\
\texttt{lens\_\allowbreak recommendation}: & \texttt{0} or \texttt{3} \\
\texttt{top\_\allowbreak k\_\allowbreak hint}: & \texttt{3}, \texttt{5}, or \texttt{8} \\
\texttt{prefer\_\allowbreak recent}: & \texttt{0} or \texttt{1} \\
\texttt{retrieval\_\allowbreak scope}: & \texttt{LOCAL} or \texttt{GLOBAL} \\
\texttt{lens\_\allowbreak confidence}: & confidence score for the selected lens \\
\texttt{lens\_\allowbreak reason}: & concise reason for the selected lens \\
\end{tabularx}
\renewcommand{\arraystretch}{1.0}

\vspace{0.7em}
\noindent\textbf{Anonymized example output:}
\begin{quote}
\scriptsize
\ttfamily
\begin{adjustbox}{max width=\linewidth}
\begin{tabular}{@{}l@{}}
\{ \\
\quad "action": "MATCH", \\
\quad "query\_type": "DYNAMIC", \\
\quad "phase": "REVISE", \\
\quad "taxonomy": \{ \\
\qquad "need\_memory": 1, \\
\qquad "memory\_span": "ACTIVE\_THREAD", \\
\qquad "ambiguity": "LOW" \\
\quad \}, \\
\quad "lens\_recommendation": 3, \\
\quad "top\_k\_hint": 5, \\
\quad "prefer\_recent": 1, \\
\quad "retrieval\_scope": "LOCAL", \\
\quad "lens\_confidence": 0.82, \\
\quad "lens\_reason": "Revision of the active artifact." \\
\} \\
\end{tabular}
\end{adjustbox}
\end{quote}

\vspace{0.5em}
\noindent\textbf{Implementation note:} invalid fields are clamped during post-processing, and revision-like continuation turns are conservatively mapped to \texttt{MATCH} with memory enabled.
\end{minipage}%
}
\caption{QTM structured output contract used by the \realm-mode implementation, with an anonymized example output.}
\label{fig:qtm-output-template}
\end{figure}

\FloatBarrier
\subsection{QTM Prompt Template}
\label{app:qtm_prompt}

The \qtm prompt asks the controller to infer decision-relevant session state without generating a final answer. In \realm mode, the prompt has three input blocks: \texttt{PREVIOUS ACTIVE CONTEXT}, \texttt{NEW USER INPUT}, and \texttt{SYSTEM RULES}. It requires strict JSON output matching Table~\ref{tab:qtm-schema}. Controller outputs are post-processed using the fixed schema and conservative continuation rules. Invalid fields are clamped to allowed values, and parsing failures use conservative fallback state choices, including continuation with memory enabled for revision-like turns.

\begin{figure}[htbp]
\centering
\fbox{%
\begin{minipage}{0.96\linewidth}
\small
\textbf{QTM prompt excerpt used in \realm mode.}

\vspace{0.6em}
\noindent\textbf{Role instruction:} You are a lightweight QTM for RealMem-style long-term multi-session interactions. Decide whether the new turn continues the active thread or starts a new topic. Classify query type and phase. Return JSON only.

\vspace{0.6em}
\noindent\textbf{Inputs provided to QTM:}
\begin{itemize}[leftmargin=*,topsep=2pt,itemsep=1pt]
    \item \texttt{PREVIOUS ACTIVE CONTEXT}: active-thread state and recent relevant context.
    \item \texttt{NEW USER INPUT}: the current user query.
    \item \texttt{SYSTEM RULES}: allowed values and conservative memory/split rules.
\end{itemize}

\vspace{0.6em}
\noindent\textbf{Required output:} strict JSON with \texttt{action}, \texttt{query\_\allowbreak type}, \texttt{phase}, \texttt{taxonomy}, \texttt{lens\_\allowbreak recommendation}, \texttt{top\_\allowbreak k\_\allowbreak hint}, \texttt{prefer\_\allowbreak recent}, and \texttt{retrieval\_\allowbreak scope}. Allowed values are those listed in Table~\ref{tab:qtm-schema}.

\vspace{0.6em}
\noindent\textbf{Failure / ambiguity rule:} represent uncertainty through conservative schema choices rather than inventing hidden state; when revision or finalization evidence is present, prefer continuation with memory enabled unless the turn clearly starts a new topic.
\end{minipage}}
\caption{Compact QTM prompt excerpt used by the implementation.}
\label{fig:qtm-prompt-template}
\end{figure}

\FloatBarrier
\subsection{Memory Construction and Prompt Assembly}
\label{app:prompt_assembly}

Given the QTM state, \method first derives memory-view controls $m_t=\psi_{\mathrm{mem}}(z_t)$ and then constructs a pre-packing candidate view from active-thread state, recent turns, and retrieved historical memory:
\begin{equation}
    \tilde c_t =
    \operatorname{Select}_{m_t}
    \left(
    s_t^{\mathrm{active}},
    h_t^{\mathrm{recent}},
    r_t^{\mathrm{hist}}
    \right).
\end{equation}
Here $m_t$ controls which blocks are included, how much historical evidence is retrieved, and their ordering in the candidate view. Model-specific packing is performed after routing. This implements the distinction between access and exposure: a large memory store may be available, but only the decision-relevant memory view should be placed in the generator prompt.

\begin{table}[htbp]
\centering
\small
\setlength{\tabcolsep}{5pt}
\renewcommand{\arraystretch}{1.12}
\begin{tabularx}{\textwidth}{@{}P{0.21\textwidth}P{0.14\textwidth}P{0.14\textwidth}P{0.17\textwidth}Y@{}}
\toprule
QTM condition & Active state & Recent turns & Historical retrieval & Intended behavior \\
\midrule
\texttt{taxonomy.\allowbreak need\_\allowbreak \allowbreak memory=0} or \texttt{lens\_\allowbreak \allowbreak recommendation=0} & omitted or minimal & optional short window & none & Avoid unnecessary context expansion for self-contained turns. \\
\texttt{action=MATCH}, \texttt{retrieval\_\allowbreak scope=LOCAL} & included & included & none or active-thread artifacts & Preserve the active thread for continuation, local edit, or verification turns. \\
\texttt{phase=REVISE} or \texttt{phase=VERIFY} & included & included & optional local evidence & Support revision and verification without drifting from the current artifact. \\
\texttt{retrieval\_\allowbreak \allowbreak scope=GLOBAL} & optional active summary & short recency window & ranked global candidates & Recover cross-session evidence for temporal or resumed-topic queries. \\
\texttt{prefer\_\allowbreak recent=1} & included if active & recency-biased & recent candidates prioritized & Bias memory construction toward recent history when recency is decision-relevant. \\
\texttt{lens\_\allowbreak recommendation=3} or \texttt{top\_\allowbreak k\_\allowbreak hint=8} & included & included & expanded ranked evidence & Provide a richer memory view for synthesis, finalization, or high-difficulty turns. \\
\bottomrule
\end{tabularx}
\caption{Memory-view construction policy. QTM emits top-$k$ hints in \{3,5,8\} for the main \method policy. The three-persona ablation slice uses a matched top-10 historical retrieval budget across the main memory-construction variants to control exposure budget.}
\label{tab:memory-view-policy}
\end{table}

After the bandit selects $a_t$, the candidate view $\tilde c_t$ is packed into $c_t(a_t)$ using that model's context window, output reservation, and safety margin. The final generator prompt has four ordered blocks: \texttt{SYSTEM RULES}, \texttt{MEMORY GUIDANCE}, \texttt{MEMORY}, and \texttt{USER QUERY}. The guidance contains the QTM-derived state header but omits internal controller metadata. The total input budget is
\begin{equation}
    B_{\mathrm{in}}(a_t)=W_{\mathrm{ctx}}(a_t)-R_{\mathrm{out}}(a_t)-M_{\mathrm{safe}}(a_t),
\end{equation}
where $W_{\mathrm{ctx}}$ is the model context window, $R_{\mathrm{out}}$ the reserved output budget, and $M_{\mathrm{safe}}$ a safety margin. The memory blocks are clipped to fit the input budget together with the system rules, guidance, and query. Token lengths are estimated using the release implementation's model-specific budget rules rather than computed exactly with each model's tokenizer. A safety fallback to another model triggers fresh packing of the candidate view under the fallback model's budget. Both benchmark runners follow this candidate-view, routing, and model-specific packing sequence.

\begin{figure}[htbp]
\centering
\fbox{%
\begin{minipage}{0.94\linewidth}
\small
\textbf{Generator prompt assembly used by \method.}

\vspace{0.7em}
\noindent\textbf{1. \texttt{SYSTEM RULES}:} global assistant and task constraints.

\vspace{0.7em}
\noindent\textbf{2. \texttt{MEMORY GUIDANCE}:} QTM-derived memory guidance, including continuation/topic-shift state, query type, phase, retrieval scope, recency preference, and lens recommendation. Internal controller scores and router uncertainty terms are not exposed.

\vspace{0.7em}
\noindent\textbf{3. \texttt{MEMORY}:} ranked candidate memory blocks, packed and clipped after model selection using the selected model's budget.

\vspace{0.7em}
\noindent\textbf{4. \texttt{USER QUERY}:} the current user request.
\end{minipage}}
\caption{Generator prompt assembly used by \method.}
\label{fig:generator-prompt-template}
\end{figure}

\FloatBarrier
\subsection{Online Router Features and Update}
\label{app:router_features}

The online router receives a feature vector
\begin{equation}
    x_t = \phi(z_t, \tilde c_t, q_t),
\end{equation}
where $\phi$ encodes QTM fields, memory-view indicators, retrieval-scope signals, interaction phase, turn depth, and pre-packing candidate-load statistics. Features are extracted before selecting the generator and do not use the final model-specific packed context. We do not include raw topic identity as a categorical feature.

\begin{table}[htbp]
\centering
\small
\setlength{\tabcolsep}{5pt}
\renewcommand{\arraystretch}{1.12}
\begin{tabularx}{\textwidth}{@{}P{0.20\textwidth}YP{0.18\textwidth}Y@{}}
\toprule
Feature group & Examples & Encoding & Purpose \\
\midrule
QTM categorical state & topic relation, query type, memory need, retrieval scope, memory view, interaction phase & one-hot / categorical & Main session-state signal for model selection. \\
Memory-load features & candidate memory blocks, retrieved-history flag, estimated candidate-token count & normalized numeric & Estimates the load of the pre-packing candidate view. \\
Session-position features & turn depth, recent revision flag, finalization flag & numeric / binary & Distinguishes early continuation from later synthesis or verification. \\
Retrieval features & retrieval scope, number of candidates, retrieval score summaries & numeric / categorical & Indicates whether the response depends on recovered evidence. \\
Cost features & model-specific cost enters through reward rather than direct decision-time subtraction & reward-side scalar & Lets the router learn quality--cost utility from observed outcomes. \\
\bottomrule
\end{tabularx}
\caption{Router feature groups. The implementation uses structured state and pre-packing candidate statistics rather than raw topic identifiers.}
\label{tab:router-features}
\end{table}

We instantiate the online router with a LinUCB-style contextual bandit:
\begin{equation}
    a_t = \arg\max_{a \in \mathcal{A}}
    \left(
    \hat{\theta}_a^\top x_t
    + \alpha \sqrt{x_t^\top A_a^{-1}x_t}
    \right).
\end{equation}
The router is updated using the turn-level reward:
\begin{equation}
    \pi_{t+1} = \operatorname{Update}(\pi_t, x_t, a_t, \hat{r}_t).
\end{equation}
LinUCB is used because it supports lightweight online updates over structured features. Other contextual bandit policies could be substituted without changing the \method interface.

\begin{table}[htbp]
\centering
\small
\setlength{\tabcolsep}{5pt}
\renewcommand{\arraystretch}{1.08}
\begin{tabularx}{\textwidth}{@{}P{0.26\linewidth}Y@{}}
\toprule
Hyperparameter / setting & Final value \\
\midrule
Candidate model pool $\mathcal{A}$ & Main \realm runs: Llama-3.2-3B, Mixtral-8x7B, Qwen-72B, GPT-4o; three-persona ablation pool: Llama3B, Qwen2.5-72B, GPT-4o. \\
Feature dimension & 30 \\
LinUCB exploration coefficient $\alpha$ & 0.8 \\
Regularization / initialization & per-arm $A_a=\lambda_{\mathrm{reg}}I$ with $\lambda_{\mathrm{reg}}=1.0$ and $b_a=0$ \\
Feature normalization & categorical one-hot plus normalized numeric features, including retrieval-budget, similarity, retrieval-count, and binary state flags \\
Warm-start policy & cold-start online policy for \method; the no-update ablation freezes this cold-start policy without online updates \\
Reward cost weight & $\lambda_{\mathrm{cost}}=0.002$ in the main \realm full-slice router run; $\lambda_{\mathrm{cost}}=0.5$ in the three-persona \realm ablation \\
\bottomrule
\end{tabularx}
\caption{Router hyperparameters and settings.}
\label{tab:router-hparams}
\end{table}

The main \realm run and the three-persona ablation use different cost weights, candidate pools, and retrieval budgets. They are separate experimental configurations rather than a single hyperparameter sweep. All component-ablation variants use $\lambda_{\mathrm{cost}}=0.5$, so their comparisons remain within the diagnostic configuration.

\FloatBarrier
\subsection{Turn-Level Judge and Reward}
\label{app:turn_judge}

During online adaptation, \method uses a turn-level judge to estimate local reward for the selected model response. In our experiments, this online feedback judge is implemented with GPT-4o-mini and is used only to update the routing policy; it is not used for benchmark reporting.

The turn-level judge receives the current query, final packed memory context $c_t(a_t)$, QTM state, generated answer, and non-final current-turn diagnostics when available. It does not receive future turns, final labels, gold final answers, or benchmark-level pass decisions. This separation prevents online updates from using information that would be unavailable at decision time.

\paragraph{Reward aggregation.}
The six rubric scores in Table~\ref{tab:turn-judge-rubric} are normalized by dividing by 5. Benchmark-specific mappings combine these scores with coverage signals and failure indicators into three components: memory use $M_t$, answer correctness $A_t$, and answer quality $Q_t$.

Both benchmarks use
\begin{equation}
\begin{aligned}
    \hat{u}_t &= \operatorname{clip}\!\bigl(0.45M_t+0.40A_t \\
              &\qquad {}+0.15Q_t-P_t,\ 0,\ 1\bigr), \\
    \hat{r}_t &= \hat{u}_t-\lambda C_t^{\mathrm{gen}}(a_t,c_t(a_t)).
\end{aligned}
\label{eq:judge-composite-reward}
\end{equation}
Here $\hat{u}_t=J_{\mathrm{turn}}(y_t,q_t,z_t,c_t(a_t))$ is the pre-cost utility defined in the main text. The weights $0.45$, $0.40$, and $0.15$ apply to the three composite components, not directly to the six raw rubric scores.

For \realm, $M_t$ averages retrieval coverage, normalized state alignment, and normalized routing adequacy. The answer component $A_t$ averages normalized helpfulness, normalized state alignment, and one minus the hallucination flag. The quality component $Q_t$ averages normalized intermediate-product and final-integration scores, using the intermediate-product score alone on non-final steps. For \mc, the memory component additionally includes normalized revision application; coverage is set to 1 when explicit memory supervision is unavailable.

\paragraph{Failure penalties.}
With binary flags ordered as topic drift, restart, hallucination, and severe missing history, $P_t$ is their weighted sum with coefficients $(0.10,0.06,0.12,0.12)$ for \realm and $(0.10,0.12,0.10,0.10)$ for \mc. The routing reward subtracts only the selected generator's cost; QTM charges remain part of the reported API cost. The released implementation will include the prompts, benchmark-specific component mappings, and reward-processing configurations used in the experiments.

\begin{table}[htbp]
\centering
\small
\setlength{\tabcolsep}{5pt}
\renewcommand{\arraystretch}{1.12}
\begin{tabularx}{\textwidth}{@{}P{0.23\textwidth}YP{0.12\textwidth}P{0.18\textwidth}@{}}
\toprule
Judge output & Definition & Range & Scoring focus \\
\midrule
\texttt{answer\_\allowbreak helpfulness} & Helpfulness and directness of the candidate answer for the current query. & 0--5 & helpfulness \\
\texttt{state\_\allowbreak alignment} & Consistency with active topic, session state, constraints, and memory context. & 0--5 & fidelity diagnostic \\
\texttt{revision\_\allowbreak application} & Whether revision instructions are applied without restarting, overwriting, or drifting. & 0--5 & revision diagnostic \\
\texttt{intermediate\_\allowbreak product} & Quality of intermediate artifacts or state updates when the turn is not final. & 0--5 & intermediate quality \\
\texttt{routing\_\allowbreak adequacy} & Whether the selected model is adequate for the state and context. & 0--5 & routing diagnostic \\
\texttt{final\_\allowbreak integration} & Whether the response integrates prior state and evidence into the final answer. & 0--5 & fidelity diagnostic \\
Hard-failure flags & Topic drift, restart instead of revision, hallucinated context use, or severe missing-history use. & 0/1 & penalty \\
\bottomrule
\end{tabularx}
\caption{Turn-level judge outputs. Normalized rubric scores, coverage signals, and failure indicators form the three composite components in Eq.~\ref{eq:judge-composite-reward}, weighted by 0.45, 0.40, and 0.15. A benchmark-specific failure penalty is applied before clipping; generation cost is then subtracted.}
\label{tab:turn-judge-rubric}
\end{table}

\begin{figure}[htbp]
\centering
\fbox{%
\begin{minipage}{0.94\linewidth}
\small
\textbf{Turn-level judge prompt excerpt.}

\vspace{0.8em}
\noindent\textbf{Inputs:}
\begin{itemize}[leftmargin=*,topsep=2pt,itemsep=1pt]
    \item Current query, injected memory context, and candidate answer.
    \item Optional non-final current-turn diagnostics and optional router metadata; no gold final answers or benchmark pass labels.
    \item Deterministic diagnostic signals, such as retrieval/context coverage and intermediate-state signals.
    \item Mode note: \texttt{synthetic}, \texttt{realmem}, or \texttt{mc\_\allowbreak router}.
\end{itemize}

\vspace{0.6em}
\noindent\textbf{Scoring instruction:} act as a strict but calibrated evaluator for step-level routing and memory use. Score dimensions A--F on a 0--5 scale, set hard-failure flags, and use deterministic auto-signals as evidence unless contradicted by the text.

\vspace{0.6em}
\noindent\textbf{Required output:} JSON only, including six scores, hard-failure flags, bounded reward information, and confidence values.
\end{minipage}}
\caption{Compact turn-level judge prompt excerpt used for online feedback.}
\label{fig:turn-judge-prompt-template}
\end{figure}

\FloatBarrier
\subsection{Cost Accounting}
\label{app:cost_accounting}

For \realm, the main-table API cost is
\begin{equation}
C_{\mathrm{gen+ctrl}}=\sum_t(C_t^{\mathrm{gen}}+C_t^{\mathrm{ctrl}}).
\end{equation}
Online feedback and final benchmark evaluators are excluded from this main-table view. Local memory operations have no external API charge in our setup, but their compute, storage, latency, and engineering costs are unpriced. The online reward subtracts only generator cost because the router chooses the generator.

On \realm, generation-controller API costs are \$26.6480 for \method, \$184.3967 for full-context GPT-4o, and \$20.2875 for A-MEM+MetaLLM. The \method/GPT-4o cost ratio is 14.5\%.

The feedback-inclusive totals in Section~\ref{sec:results-realmem} are estimated from logged online-judge input-token counts and conservatively estimated output usage, priced at \$0.15 and \$0.60 per million tokens, respectively. This adds approximately \$5.88 for \method and \$3.20 for A-MEM+MetaLLM, giving approximately \$32.53 and \$23.49. These estimates exclude final evaluators and local compute. For \mc, costs follow the original run-level recording convention, without a separate QTM/controller breakdown.

\begin{table}[htbp]
\centering
\small
\setlength{\tabcolsep}{5pt}
\begin{tabularx}{\textwidth}{@{}P{0.20\linewidth}P{0.34\linewidth}YY@{}}
\toprule
Component & Role / model & Input price & Output price \\
\midrule
QTM/controller & \texttt{qwen/qwen-2.5-7b-instruct} & \$0.04 / 1M tokens & \$0.10 / 1M tokens \\
Turn-level judge & GPT-4o-mini online feedback judge & \$0.15 / 1M tokens & \$0.60 / 1M tokens \\
Final \realm evaluator & GPT-4o final benchmark judge where required & excluded from both views & excluded \\
Final \mc evaluator & GPT-4o path-aware final judge & excluded from both views & excluded \\
Generators & candidate pool $\mathcal{A}$ & Llama3B: \$0.05; Mixtral: \$0.20; Qwen72B: \$0.35; GPT-4o: \$2.50 / 1M tokens & Llama3B: \$0.10; Mixtral: \$0.20; Qwen72B: \$0.40; GPT-4o: \$10.00 / 1M tokens \\
Memory search/add & \textsc{Mem0}/\textsc{A-MEM} where applicable & no external embedding API; local compute unpriced & same \\
\bottomrule
\end{tabularx}
\caption{Cost-accounting components and the unit prices used for the reported runs. Prices are per 1M tokens where applicable.}
\label{tab:cost-prices}
\end{table}

\FloatBarrier
\subsection{Uncertainty and Interaction-Order Diagnostics}
\label{app:uncertainty_order}

Empirical bootstrap resampling over the 10 \realm personas gives 95\% intervals of [0.7183, 0.7641] for QA, [0.6571, 0.7024] for Mem Recall, and [\$2.4004, \$2.9394] for per-persona generation-controller cost. The corresponding persona-macro centers are 0.7406, 0.6794, and \$2.6648, matching the aggregation used in the main table. As an aggregation check, direct query-weighted \method values are 0.7404 QA over 1,411 scored queries and 0.6815 Mem Recall over 1,393 scored queries; full-context GPT-4o obtains 0.7529 and 0.1655, respectively, over 1,414 scored queries. The substantive comparison is unchanged. Scenario-level empirical bootstrap gives a full-\mc pass rate of
0.2367 with a 95\% interval of [0.1837, 0.2898] and a focused-subset
rate of 0.3212 with a 95\% interval of [0.2409, 0.4015], conditional
on the evaluated order.

In \realm, each persona has an independently initialized router. Reordering personas therefore does not alter their router updates, while within-persona chronology defines the causal history and is preserved. The bootstrap intervals quantify sample variation across personas or scenarios, conditional on the evaluated histories and ordering.

For a separate interaction-order diagnostic, we replay the same fixed
30-scenario \mc subset under eight randomized chunk orders while holding
the evaluated cases and all non-order configurations fixed. The original
scenario sequence is partitioned into ten chunks of three scenarios; each
seed shuffles only the chunk order, with within-chunk order preserved.
Each replay starts from a freshly initialized bandit in a separate process,
and online updates are confined to that replay. We retain the main
\mc routing configuration, including chunk size 3, chunk-level memory
scope, Qwen2.5-7B as QTM, $\lambda_{\mathrm{cost}}=0.002$, the same
four-model router pool, and unchanged router features and decoding settings.

To keep the order comparison internally consistent, all eight replays are
evaluated with the same fixed GPT-4o-mini path-aware evaluator. The primary
\mc benchmark results use the GPT-4o final evaluator described in
Appendix~\ref{app:final_judges}.

\begin{table}[htbp]
\centering
\footnotesize
\setlength{\tabcolsep}{5pt}
\renewcommand{\arraystretch}{1.05}
\begin{adjustbox}{max width=\linewidth}
\begin{tabular}{@{}rcc@{}}
\toprule
Seed & Final pass & Pass rate \\
\midrule
13  & 5/30 & 16.7\% \\
29  & 4/30 & 13.3\% \\
47  & 7/30 & 23.3\% \\
61  & 6/30 & 20.0\% \\
79  & 7/30 & 23.3\% \\
97  & 5/30 & 16.7\% \\
113 & 5/30 & 16.7\% \\
131 & 5/30 & 16.7\% \\
\midrule
Order mean & 5.5/30 & 18.3\% \\
\bottomrule
\end{tabular}
\end{adjustbox}
\caption{Interaction-order diagnostic on the same fixed set of 30
\mc scenarios across eight randomized chunk orders. The sample standard
deviation of the order-level pass rates is 3.6 percentage points.}
\label{tab:mc-order-diagnostic}
\end{table}

Across the eight sampled orders, pass counts range from 4/30 to 7/30,
with a mean pass rate of 18.3\% and a sample standard deviation of
3.6 percentage points. This diagnostic characterizes variation across
the tested processing orders rather than absolute benchmark performance.

\section{Benchmark Details}
\label{app:benchmark_details}

\FloatBarrier
\subsection{\realm Protocol and Metric Normalization}
\label{app:realmem_protocol}

\realm evaluates long-term memory-driven interaction~\citep{bian2026realmem}. We use the evaluation pool of 10 personas and 1,415 queries and report normalized QA, Mem Recall, Recall@20, NDCG@20, and API cost. Main-table quality and memory metrics are macro-averaged over the 10 personas; Section~\ref{app:uncertainty_order} reports direct query-weighted aggregates and their scored coverage separately.

The original \realm QA judge score is reported on a 0--3 scale. Following the normalized metric used in the main text, we report
\begin{equation}
    \mathrm{QA}_{\mathrm{norm}} = \frac{\mathrm{QA}_{\mathrm{raw}}}{3}.
\end{equation}
All \realm QA values in Appendix~\ref{app:realmem-full} use this normalized score, so they align with Table~\ref{tab:realmem-main} in the main text. Mem Recall is the \realm memory judge score. Recall@20 and NDCG@20 measure retrieval coverage and ranking quality.

Recall@20 and NDCG@20 are taken directly from the official retrieval evaluator, without redefining the metrics for full-context baselines. They assess the retrieval output scored by that pipeline; the full-context label describes the generator input and does not imply a retrieval score of one. Fixed generators paired with the same retrieval output therefore share retrieval scores even when their QA differs.

\FloatBarrier
\subsection{\mc Conversion and Axes}
\label{app:mc_protocol}

Of the 273 official \mc scenarios, we reserved 28 (approximately 10\%)
for QTM prompt development and preliminary testing and excluded them
from the reported evaluation split. The remaining 245 scenarios,
comprising 1,480 turns, form the evaluation split; the split identifiers
will be released with the implementation. We convert each scenario into a session-level memory-routing trajectory: intermediate turns provide the path over which systems must maintain state. The final verification turn uses the scenario's official binary rubric question as the last user query. Under the supplied pass criterion, a trajectory satisfying the required memory and constraints supports the expected answer \texttt{YES}. This question originates from the official evaluation rubric; placing it in the assistant's input trajectory is part of our conversion.

In the converted evaluation format, intermediate turns are marked as \texttt{eval\_\allowbreak type=0}; the final official rubric question is marked as \texttt{eval\_\allowbreak type=1} with its supplied \texttt{PASS\_\allowbreak CRITERIA}; exactly one \texttt{eval\_\allowbreak type=1} appears per scenario and it is always the final turn. The benchmark contains four disjoint axes:
\begin{itemize}[leftmargin=*,topsep=2pt,itemsep=1pt]
    \item \textsc{Inference Memory}: 102 scenarios.
    \item \textsc{Instruction Retention}: 62 scenarios.
    \item \textsc{Self Coherence}: 46 scenarios.
    \item \textsc{Reliable Version Editing}: 35 scenarios.
\end{itemize}
The main text reports the full 245-scenario aggregate first, then analyzes \textsc{Inference Memory} and \textsc{Reliable Version Editing} (137 scenarios) as a mechanism-focused subset. The per-axis table is in Appendix~\ref{app:mc-axis-breakdown}.

\FloatBarrier
\subsection{Chunked Context Construction}
\label{app:chunked_context}

The current \mc construction uses chunk size 3. A chunk bundles several session trajectories into a longer context block, creating topic interference. The system must identify the active trajectory, select relevant evidence, and ignore distractor sessions in the same chunk.

The construction procedure is:
\begin{enumerate}[leftmargin=*,topsep=2pt,itemsep=1pt]
    \item Group scenarios into chunks of size 3.
    \item For each chunk, concatenate the trajectories into a shared context block.
    \item Mark one scenario trajectory as the active trajectory for the current query.
    \item Present the system with the chunked context and the active final-turn request.
    \item Evaluate whether the final answer follows the active trajectory rather than distractor trajectories.
\end{enumerate}
This setting stresses both retrieval selectivity and model capacity because the correct response requires maintaining the active state under structured interference.

\FloatBarrier
\subsection{Final Evaluation Judges}
\label{app:final_judges}

For \realm, we use the official retrieval and generation evaluation pipeline. The final reporting protocol uses GPT-4o as the generation/memory judge model where required, with retrieval-memory helpfulness evaluated using top-$k=5$ evidence. This final evaluator is used only for benchmark reporting and is distinct from the step-level feedback path used for online routing updates.

For \mc, we use a path-aware final judge following the benchmark's evaluation style. The judge first considers the trajectory through turns $1$ to $t_{\max}-1$: whether the system maintained the correct topic memory, preserved constraints, and produced intermediate responses that support the final answer. It then evaluates the final turn at $t_{\max}$ by asking whether the maintained memory supports the response and whether the response itself is correct. A scenario is counted as passed under strict joint logic: the continuity/path verdict, final-answer judge verdict, and parseable model final-token answer must all satisfy the supplied pass criterion, with \texttt{YES} as the expected answer to the final official rubric question.

\begin{figure}[htbp]
\centering
\fbox{%
\begin{minipage}{0.94\linewidth}
\small
\textbf{Path-aware \mc final judge protocol.}

\vspace{0.8em}
\noindent\textbf{Inputs:}
\begin{itemize}[leftmargin=*,topsep=2pt,itemsep=1pt]
    \item Full trajectory through turns $1$ to $t_{\max}-1$.
    \item The scenario's official binary rubric question, used as the final user request at $t_{\max}$ with ``Please answer with only YES or NO.''
    \item System final answer.
    \item Scenario-specific \texttt{PASS\_\allowbreak CRITERIA} or expected behavior.
\end{itemize}

\vspace{0.6em}
\noindent\textbf{Strict joint pass logic:} a scenario passes only when the judge accepts memory-continuity/path consistency, accepts final-answer correctness, and the model's final-token answer is parseable and matches the expected \texttt{YES} under the supplied pass criterion.

\vspace{0.6em}
\noindent\textbf{Evaluator setting:} GPT-4o with deterministic judge decoding, temperature 0, and a short max-token budget.
\end{minipage}}
\caption{Path-aware \mc final judging protocol used for the fixed evaluation matrix.}
\label{fig:mc-final-judge-prompt}
\end{figure}

\section{Full Results}
\label{app:additional_results}
\label{app:full_results}

This section reports the full benchmark matrices. The three-persona component ablations are presented separately in Appendix~\ref{app:ablations}.

\FloatBarrier
\subsection{\realm Full Matrix}
\label{app:realmem-full}

Table~\ref{tab:realmem-full} reports the full matrix. Context-limit failures are identified from provider error payloads: 1,344 Qwen2.5-72B queries, 1,344 Mixtral queries, and 1,247 Llama3B queries. Affected queries receive zero score and zero recorded cost.

\begin{table}[htbp]
\centering
\small
\setlength{\tabcolsep}{5pt}
\renewcommand{\arraystretch}{1.06}
\begin{adjustbox}{max width=\linewidth}
\begin{tabular}{@{}llrrrrr@{}}
\toprule
Memory Setting & Model / Router & Gen.-Ctrl. (\$) $\downarrow$ & Norm. QA $\uparrow$ & Mem Rec. $\uparrow$ & R@20 $\uparrow$ & NDCG@20 $\uparrow$ \\
\midrule
\multicolumn{2}{c}{\textbf{\method}} & 26.6480 & \underline{0.7406} & \textbf{0.6794} & \textbf{0.8015} & \textbf{0.6168} \\
\midrule
Full context & GPT-4o & 184.3967 & \textbf{0.7540} & 0.1640 & 0.5188 & 0.3216 \\
Full context & \textsc{MetaLLM} & 180.0189 & 0.7331 & 0.1603 & 0.5188 & 0.3216 \\
Full context & Qwen2.5 72B$^\dagger$ & 0.4737 & 0.0422 & 0.2041 & 0.5188 & 0.3216 \\
Full context & Mixtral$^\dagger$ & 0.2970 & 0.0336 & 0.1577 & 0.5188 & 0.3216 \\
Full context & Llama3B$^\dagger$ & 0.2911 & 0.0214 & 0.1605 & 0.5188 & 0.3216 \\
\textsc{Mem0} & GPT-4o & 41.3834 & 0.6909 & 0.5997 & 0.6305 & 0.5388 \\
\textsc{Mem0} & \textsc{MetaLLM} & 24.3499 & 0.6652 & 0.6048 & 0.6305 & 0.5388 \\
\textsc{Mem0} & Qwen2.5 72B & 3.8325 & 0.5137 & 0.5961 & 0.6305 & 0.5388 \\
\textsc{Mem0} & Mixtral & 3.5789 & 0.5675 & 0.6035 & 0.6305 & 0.5388 \\
\textsc{Mem0} & Llama3B & 0.9383 & 0.2689 & 0.6059 & 0.6305 & 0.5388 \\
\textsc{A-MEM} & GPT-4o & 43.1311 & 0.7172 & \underline{0.6479} & \underline{0.6817} & \underline{0.5846} \\
\textsc{A-MEM} & \textsc{MetaLLM} & 20.2875 & 0.7044 & 0.6471 & \underline{0.6817} & \underline{0.5846} \\
\textsc{A-MEM} & Qwen2.5 72B & 3.5106 & 0.4612 & 0.6447 & \underline{0.6817} & \underline{0.5846} \\
\textsc{A-MEM} & Mixtral & 3.7563 & 0.5796 & 0.6445 & \underline{0.6817} & \underline{0.5846} \\
\textsc{A-MEM} & Llama3B & 1.0175 & 0.2891 & 0.6475 & \underline{0.6817} & \underline{0.5846} \\
\bottomrule
\end{tabular}
\end{adjustbox}
\caption{Full \realm matrix. Quality and memory metrics are persona-macro averages; cost records generation and controller API calls. $^\dagger$Context-limit failures affect 1,344/1,415 Qwen queries, 1,344/1,415 Mixtral queries, and 1,247/1,415 Llama3B queries, so scores conflate executability and response quality.}
\label{tab:realmem-full}
\end{table}

\FloatBarrier
\subsection{\mc Full Aggregate}
\label{app:mc-full}

\begin{table}[htbp]
\centering
\small
\setlength{\tabcolsep}{5pt}
\renewcommand{\arraystretch}{1.06}
\begin{adjustbox}{max width=\linewidth}
\begin{tabular}{@{}llrrrr@{}}
\toprule
Memory Setting & Model / Router & Pass Rate $\uparrow$ & Pass $\uparrow$ & Recorded Cost (\$) $\downarrow$ & Cost/Pass (\$) $\downarrow$ \\
\midrule
\multicolumn{2}{c}{\textbf{\method}} & 0.2367 & 58/245 & 3.2504 & 0.0560 \\
\midrule
Full context & Qwen2.5 72B & \textbf{0.2735} & \textbf{67/245} & 3.1665 & 0.0473 \\
Full context & GPT-4o & \underline{0.2612} & \underline{64/245} & 15.3491 & 0.2398 \\
\textsc{Mem0} & Mixtral & 0.2327 & 57/245 & 0.7712 & 0.0135 \\
\textsc{Mem0} & GPT-4o & 0.2286 & 56/245 & 13.9769 & 0.2496 \\
\textsc{Mem0} & \textsc{MetaLLM} & 0.2122 & 52/245 & 4.7825 & 0.0920 \\
Full context & \textsc{MetaLLM} & 0.2082 & 51/245 & 6.4945 & 0.1273 \\
Full context & Mixtral & 0.1918 & 47/245 & 1.2327 & 0.0262 \\
\textsc{Mem0} & Qwen2.5 72B & 0.1918 & 47/245 & 2.5946 & 0.0552 \\
\textsc{A-MEM} & \textsc{MetaLLM} & 0.1837 & 45/245 & 4.7664 & 0.1059 \\
\textsc{A-MEM} & GPT-4o & 0.1796 & 44/245 & 11.3333 & 0.2576 \\
\textsc{A-MEM} & Qwen2.5 72B & 0.1796 & 44/245 & 1.4152 & 0.0322 \\
\textsc{A-MEM} & Mixtral & 0.1673 & 41/245 & 0.7823 & 0.0191 \\
Full context & Llama3B & 0.0776 & 19/245 & 0.6086 & 0.0320 \\
\textsc{Mem0} & Llama3B & 0.0694 & 17/245 & 0.5448 & 0.0320 \\
\textsc{A-MEM} & Llama3B & 0.0571 & 14/245 & 0.2617 & 0.0187 \\
\bottomrule
\end{tabular}
\end{adjustbox}
\caption{\mc full aggregate over 245 scenarios, the primary benchmark view. Costs follow the original run-level accounting convention; QTM/controller cost was not separately itemized in the archived logs.}
\label{tab:mc-full}
\end{table}

\FloatBarrier
\subsection{\mc Mechanism-Focused Subset}
\label{app:mc-target}

Table~\ref{tab:mc-target-appendix} reports all 16 systems on the mechanism-focused subset. It supplements rather than replaces the full aggregate.

\begin{table}[htbp]
\centering
\small
\setlength{\tabcolsep}{5pt}
\renewcommand{\arraystretch}{1.06}
\begin{adjustbox}{max width=\linewidth}
\begin{tabular}{@{}llrrrr@{}}
\toprule
Memory Setting & Model / Router & Pass Rate $\uparrow$ & Pass $\uparrow$ & Recorded Cost (\$) $\downarrow$ & Cost/Pass (\$) $\downarrow$ \\
\midrule
\multicolumn{2}{c}{\textbf{\method}} & \underline{0.3212} & \underline{44/137} & 1.5670 & 0.0356 \\
\midrule
Full context & Qwen2.5 72B & \textbf{0.3285} & \textbf{45/137} & 1.8528 & 0.0412 \\
\textsc{Mem0} & Mixtral & 0.2847 & 39/137 & 0.4620 & 0.0118 \\
Full context & GPT-4o & 0.2628 & 36/137 & 9.1044 & 0.2529 \\
Full context & Mixtral & 0.2555 & 35/137 & 0.7534 & 0.0215 \\
\textsc{Mem0} & GPT-4o & 0.2555 & 35/137 & 8.4375 & 0.2411 \\
\textsc{Mem0} & \textsc{MetaLLM} & 0.2555 & 35/137 & 2.8092 & 0.0803 \\
\textsc{Mem0} & Qwen2.5 72B & 0.2409 & 33/137 & 1.5597 & 0.0473 \\
Full context & \textsc{MetaLLM} & 0.2263 & 31/137 & 3.6814 & 0.1188 \\
\textsc{A-MEM} & GPT-4o & 0.2044 & 28/137 & 6.8631 & 0.2451 \\
\textsc{A-MEM} & \textsc{MetaLLM} & 0.1898 & 26/137 & 2.7960 & 0.1075 \\
\textsc{A-MEM} & Mixtral & 0.1898 & 26/137 & 0.4675 & 0.0180 \\
\textsc{A-MEM} & Qwen2.5 72B & 0.1752 & 24/137 & 0.8436 & 0.0352 \\
Full context & Llama3B & 0.1022 & 14/137 & 0.3271 & 0.0234 \\
\textsc{Mem0} & Llama3B & 0.0803 & 11/137 & 0.3173 & 0.0288 \\
\textsc{A-MEM} & Llama3B & 0.0511 & 7/137 & 0.1434 & 0.0205 \\
\bottomrule
\end{tabular}
\end{adjustbox}
\caption{\mc mechanism-focused subset over \textsc{Inference Memory} and \textsc{Reliable Version Editing}; not the primary leaderboard. Costs follow the original run-level accounting convention, without a separate historical QTM/controller itemization.}
\label{tab:mc-target-appendix}
\end{table}

\FloatBarrier
\subsection{\mc Per-Axis Breakdown}
\label{app:mc-axis-breakdown}

Table~\ref{tab:mc-axis-breakdown} reports the category-level breakdown needed to contextualize the target subset and reduce ambiguity about axis-level behavior.

\begin{table}[htbp]
\centering
\small
\setlength{\tabcolsep}{3.5pt}
\renewcommand{\arraystretch}{1.05}
\begin{adjustbox}{max width=\linewidth}
\begin{tabular}{@{}llccccc@{}}
\toprule
Memory Setting & Model / Router & Inf. Mem. & Instr. Ret. & Self Coh. & Version Edit & Total \\
 & & /102 & /62 & /46 & /35 & /245 \\
\midrule
\multicolumn{2}{c}{\textbf{\method}} & \underline{33} & 8 & 6 & \textbf{11} & 58 \\
\midrule
Full context & Qwen2.5 72B & \textbf{36} & \underline{14} & 8 & \underline{9} & \textbf{67} \\
Full context & GPT-4o & 28 & \textbf{18} & \textbf{10} & 8 & \underline{64} \\
Full context & \textsc{MetaLLM} & 26 & 12 & 8 & 5 & 51 \\
Full context & Mixtral & 28 & 6 & 6 & 7 & 47 \\
Full context & Llama3B & 10 & 5 & 0 & 4 & 19 \\
\textsc{Mem0} & GPT-4o & 27 & 11 & \textbf{10} & 8 & 56 \\
\textsc{Mem0} & \textsc{MetaLLM} & 27 & 9 & 8 & 8 & 52 \\
\textsc{Mem0} & Qwen2.5 72B & 25 & 8 & 6 & 8 & 47 \\
\textsc{Mem0} & Mixtral & 31 & 9 & \underline{9} & 8 & 57 \\
\textsc{Mem0} & Llama3B & 10 & 4 & 2 & 1 & 17 \\
\textsc{A-MEM} & GPT-4o & 25 & 11 & 5 & 3 & 44 \\
\textsc{A-MEM} & \textsc{MetaLLM} & 23 & 13 & 6 & 3 & 45 \\
\textsc{A-MEM} & Qwen2.5 72B & 22 & 12 & 8 & 2 & 44 \\
\textsc{A-MEM} & Mixtral & 22 & 7 & 8 & 4 & 41 \\
\textsc{A-MEM} & Llama3B & 6 & 5 & 2 & 1 & 14 \\
\bottomrule
\end{tabular}
\end{adjustbox}
\caption{\mc per-axis pass-count breakdown. Bold and underline mark best and second-best pass counts within each axis.}
\label{tab:mc-axis-breakdown}
\end{table}

\section{Ablations and Sensitivity Analysis}
\label{app:ablations}

This section provides component-level evidence for the mechanisms used by \method. The main benchmark tables compare complete systems on the full evaluation slices; the analyses below examine QTM-guided memory construction, QTM-derived prompt state, no-QTM routing, recency-only memory selection, online updates, cost weighting, and local QTM prompt sensitivity. The main component ablations use a three-persona subset of the \realm evaluation pool, while the prompt-sensitivity probe uses frozen snapshots from seven evaluation personas. The main result aggregates all 10 personas and 1,415 queries. The diagnostic samples are drawn from this evaluation pool.

\FloatBarrier
\subsection{\realm Three-Persona Ablation Setup}
\label{app:realmem-ablation-setup}

Table~\ref{tab:realmem-ablation-setup} summarizes the ablation setting. The three personas are randomly sampled with seed 28. All main memory-construction variants use a matched top-10 retrieval budget so that differences are not explained by exposing more historical memory. The component ablation slice uses $\lambda_{\mathrm{cost}}=0.5$ for all variants in Table~\ref{tab:realmem-ablation-official}. This differs from the main full-slice \realm run because the ablation is designed as a controlled within-slice diagnostic with stronger cost pressure; the resulting scores are compared only within the ablation slice.

\begin{table}[htbp]
\centering
\small
\setlength{\tabcolsep}{5pt}
\renewcommand{\arraystretch}{1.08}
\begin{tabularx}{\linewidth}{@{}P{0.40\linewidth}Y@{}}
\toprule
Setting & Value \\
\midrule
Evaluation subset & Three personas from the 10-persona evaluation pool \\
Persona sampling & random sample with seed 28 \\
Personas & \texttt{Ethan\_\allowbreak Hunt}, \texttt{Kenta\_\allowbreak Tanaka}, \texttt{Sarah\_\allowbreak Miller} \\
Evaluated turns & 473 \\
Historical retrieval budget & top-10 \\
Model pool & Llama3B, Qwen2.5 72B, GPT-4o \\
Reward cost weight for component ablations & $\lambda_{\mathrm{cost}}=0.5$ \\
Generation temperature & 0.0 \\
Provider fallback & enabled \\
Generation error rate & 0.0 for all variants \\
\bottomrule
\end{tabularx}
\caption{\realm three-persona ablation setup. The slice is used to test component behavior under a controlled retrieval budget.}
\label{tab:realmem-ablation-setup}
\end{table}

\paragraph{Metrics.}
We report the official \realm evaluation metrics on the ablation slice. Official QA is produced on the original 0--3 scale and normalized as $\mathrm{QA}_{\mathrm{norm}}=\mathrm{QA}_{\mathrm{raw}}/3$, matching the main text. Mem Recall and Mem Helpful are official memory-evaluation metrics. Recall@10 all and NDCG@10 measure retrieval coverage and ranking quality on the ablation slice. Total cost includes generator and QTM/controller calls where applicable, following the cost convention in Appendix~\ref{app:cost_accounting}. MemFail is the percentage of evaluated turns with memory-related hard-failure flags from the turn-level diagnostic judge. The L/Q/G column reports the percentage of turns routed to Llama3B, Qwen2.5 72B, and GPT-4o, respectively.

\FloatBarrier
\subsection{Variant Definitions}
\label{app:realmem-ablation-variants}

Table~\ref{tab:realmem-ablation-variants} defines the evaluated component variants. The ablations are organized by dependency: QTM-guided context construction, QTM state/header exposure in the generator prompt, QTM-based router features, recency-only memory selection, and online updates. The no-update variant keeps the QTM-guided memory view but freezes the router at initialization. We treat this as a diagnostic stress test of online adaptation rather than as an optimized static-router baseline.

\begin{table}[htbp]
\centering
\small
\setlength{\tabcolsep}{5pt}
\renewcommand{\arraystretch}{1.08}
\begin{tabularx}{\textwidth}{@{}P{0.23\linewidth}P{0.36\linewidth}Y@{}}
\toprule
Variant & Ablated dependency & Purpose \\
\midrule
Full \method (\texttt{mars\_\allowbreak full}) & none & Full memory-aware routing system. \\
Embedding-only semantic (\texttt{mars\_\allowbreak embedctx\_\allowbreak bm}) & Replace QTM-guided memory construction with raw-query semantic top-10 retrieval & Tests whether QTM-guided context improves over embedding-only memory. \\
No QTM state/header (\texttt{mars\_\allowbreak semctx\_\allowbreak noactive}) & Start from embedding-only semantic retrieval and remove the QTM state/header from the generator prompt & Isolates prompt-state exposure relative to the embedding-only variant; its retrieval policy is unchanged. \\
Recent-only memory (\texttt{mars\_\allowbreak recentctx\_\allowbreak bm}) & Replace QTM/semantic retrieval with recent-only top-10 memory & Tests whether recency alone is sufficient. \\
No-QTM semantic (\texttt{mars\_\allowbreak noqtm\_\allowbreak semctx}) & Disable QTM; use semantic memory and query-embedding router features & Stress-tests the framework without QTM state. \\
No-QTM recent (\texttt{mars\_\allowbreak noqtm\_\allowbreak recentctx}) & Disable QTM; use recent-only memory and query-embedding router features & Tests the weakest no-QTM recency-based setting. \\
No online update (\texttt{mars\_\allowbreak full\_\allowbreak no\_\allowbreak update}) & Keep QTM-guided memory construction but freeze the cold-start router & Diagnoses adaptation from this initialization. \\
\bottomrule
\end{tabularx}
\caption{Ablation variants on the three-persona \realm evaluation slice. All variants use deterministic generation and the same model pool; historical retrieval uses a matched top-10 budget unless otherwise noted.}
\label{tab:realmem-ablation-variants}
\end{table}

\FloatBarrier
\subsection{Official Evaluation Results}
\label{app:realmem-ablation-results}

Table~\ref{tab:realmem-ablation-official} reports the official \realm evaluation metrics and diagnostic routing statistics for the component ablation variants. Full \method obtains the strongest overall profile among the online ablation variants: it has the highest normalized QA, the strongest Mem Recall, and the best retrieval metrics under the matched top-10 budget. The no-QTM semantic variant is close in QA, showing that semantic retrieval is a strong baseline, but it has lower memory scores than full \method. The recent-only variants degrade sharply on memory and retrieval metrics, indicating that simple recency is not a reliable long-session memory strategy in this setting.

The no-update diagnostic has high memory and retrieval metrics because it retains QTM-guided context construction. However, its QA is much lower: freezing the cold-start router leaves the policy concentrated on the cheapest model, which preserves retrieval quality but substantially reduces answer quality. This row should therefore be interpreted as a cold-start online-adaptation diagnostic, not as a tuned static-router baseline.

\begin{table}[htbp]
\centering
\footnotesize
\setlength{\tabcolsep}{2.8pt}
\renewcommand{\arraystretch}{1.06}
\begin{adjustbox}{max width=\linewidth}
\begin{tabular}{@{}lrrrrrrrrr@{}}
\toprule
Variant & QA raw $\uparrow$ & QA $\uparrow$ & Mem Rec. $\uparrow$ & Mem Help. $\uparrow$ & R@10 $\uparrow$ & NDCG@10 $\uparrow$ & Cost $\downarrow$ & MemFail $\downarrow$ & L/Q/G \% \\
\midrule
Full \method & \textbf{2.2333} & \textbf{0.7444} & \textbf{0.6804} & \textbf{1.6967} & \textbf{0.6744} & \textbf{0.5974} & 5.643 & \textbf{6.34} & 9.5/61.7/28.8 \\
Embedding-only semantic & 2.1692 & 0.7231 & 0.6429 & 1.6461 & 0.6026 & 0.5162 & 5.911 & 9.30 & 9.1/58.6/32.3 \\
No QTM state/header & 2.1298 & 0.7099 & 0.6397 & 1.6575 & 0.6026 & 0.5162 & 6.338 & 10.15 & 9.1/55.6/35.3 \\
Recent-only memory & 1.8837 & 0.6279 & 0.1070 & 0.2410 & 0.2981 & 0.2299 & 5.638 & 27.27 & 8.0/61.1/30.9 \\
No-QTM semantic & 2.2162 & 0.7387 & 0.6287 & 1.6174 & 0.6026 & 0.5162 & \textbf{4.570} & 10.99 & 3.0/75.9/21.1 \\
No-QTM recent & 1.8732 & 0.6244 & 0.1118 & 0.2347 & 0.2981 & 0.2299 & 4.918 & 34.67 & 8.0/66.6/25.4 \\
No online update$^\dagger$ & 1.2664 & 0.4221 & 0.6882 & 1.7098 & 0.6744 & 0.5983 & 0.439 & 18.60 & 99.6/0.0/0.4 \\
\bottomrule
\end{tabular}
\end{adjustbox}
\caption{Official \realm evaluation results on the three-persona ablation slice, with diagnostic failure and routing statistics. QA is official raw QA divided by 3; cost is generation-controller API cost in USD. Bold marks the best online component variant. $^\dagger$The no-update row is a frozen-router diagnostic, excluded from best-online markings.}
\label{tab:realmem-ablation-official}
\end{table}

\paragraph{Interpretation.}
The ablations support three conservative conclusions. First, QTM-guided memory construction provides the strongest quality--memory operating point among the main memory-construction variants under a matched top-10 retrieval budget. Full \method improves over embedding-only semantic retrieval in QA, Mem Recall, Mem Helpful, Recall@10, NDCG@10, and MemFail. Second, semantic retrieval is a strong baseline: the no-QTM semantic variant approaches full \method in normalized QA, but its lower Mem Recall and Mem Helpful scores indicate weaker memory reliability. Third, recency-only memory selection is insufficient for \realm-style long-session use, as both recent-only variants collapse on Mem Recall, Mem Helpful, Recall@10, and NDCG@10. The no-update result further shows that good memory construction alone is insufficient when the frozen router selects an underpowered generator.

The no-QTM semantic variant is cheaper and remains competitive in QA, while the no-update diagnostic is much cheaper but has substantially lower QA. These variants occupy different points on the quality--memory--cost trade-off.

\FloatBarrier
\subsection{Cost-Weight Sensitivity}
\label{app:additional-sensitivity}
\label{app:lambda_sensitivity}

Table~\ref{tab:realmem-lambda-sensitivity} reports a cost-weight sensitivity analysis over the same three-persona ablation slice. These rows keep the full \method architecture and vary only the cost coefficient in the online reward. We report the three additional settings, $\lambda_{\mathrm{cost}}\in\{0,0.25,1.0\}$; the main component-ablation setting uses $\lambda_{\mathrm{cost}}=0.5$ and is reported in Table~\ref{tab:realmem-ablation-official}. The results show that the cost coefficient shifts the quality--memory--cost operating point rather than producing a uniformly dominant setting. Removing the cost penalty slightly improves Mem Recall but increases cost and does not improve QA over the main setting. Stronger cost pressure reduces cost while maintaining comparable retrieval scores. We therefore treat $\lambda_{\mathrm{cost}}$ as a deployment knob rather than as a universally optimal scalar.

\begin{table}[htbp]
\centering
\small
\setlength{\tabcolsep}{5pt}
\renewcommand{\arraystretch}{1.06}
\begin{adjustbox}{max width=\linewidth}
\begin{tabular}{@{}lrrrrrrr@{}}
\toprule
Variant & QA $\uparrow$ & Mem Rec. $\uparrow$ & R@10 $\uparrow$ & NDCG@10 $\uparrow$ & Cost $\downarrow$ & MemFail $\downarrow$ & L/Q/G \% \\
\midrule
Full \method, $\lambda=0$ & 0.7329 & 0.6870 & 0.6702 & 0.5952 & 5.964 & 8.25 & 7.8/60.7/31.5 \\
Full \method, $\lambda=0.25$ & 0.7209 & 0.6832 & 0.6702 & 0.5933 & 5.508 & 8.25 & 10.4/61.7/27.9 \\
Full \method, $\lambda=1.0$ & 0.7225 & 0.6859 & 0.6765 & 0.5999 & 5.329 & 6.55 & 9.9/63.8/26.2 \\
\bottomrule
\end{tabular}
\end{adjustbox}
\caption{Cost-weight sensitivity on the same three-persona \realm ablation slice. These rows keep the full \method architecture and vary only the reward cost coefficient. The main component-ablation setting uses $\lambda_{\mathrm{cost}}=0.5$ and is reported in Table~\ref{tab:realmem-ablation-official}.}
\label{tab:realmem-lambda-sensitivity}
\end{table}

\FloatBarrier
\subsection{QTM Prompt Sensitivity}
\label{app:qtm-prompt-sensitivity}

We additionally test the local stability of the QTM interface under two controlled prompt changes using 150 frozen turn snapshots from seven \realm evaluation personas. For each snapshot, the user query, active context, recent history, system rules, historical memory pool, and embeddings from
\texttt{BAAI/bge-small-en-v1.5}~\citep{xiao2024cpack} are fixed. We compare the original QTM prompt with (i) a paraphrase that preserves the decision definitions and JSON schema and (ii) an instruction-reordered variant that reverses the five \texttt{Important semantics} items while leaving the remaining text unchanged. All variants use Qwen2.5-7B-Instruct with deterministic decoding. This is a controller-level diagnostic rather than an end-to-end rerun of the main benchmark.

We report two quantities with direct downstream effects. \emph{Action agreement} is the fraction of inputs whose effective \texttt{MATCH}/\allowbreak\texttt{NEW\_\allowbreak TOPIC} decision is unchanged after the same production parser and deterministic release logic. \emph{Retrieval-source Jaccard} is computed per input over the selected historical memory-chunk IDs (up to $k=20$ before final prompt-budget packing) and then averaged; double-empty pairs receive 1 and one-sided empty pairs receive 0. The original prompt is a reference, not a gold annotation, so disagreement is not an error rate.

\begin{table}[htbp]
\centering
\footnotesize
\setlength{\tabcolsep}{3pt}
\renewcommand{\arraystretch}{1.05}
\begin{adjustbox}{max width=\linewidth}
\begin{tabular}{@{}lcccc@{}}
\toprule
Prompt variant & \multicolumn{2}{c}{All 150 inputs} & \multicolumn{2}{c}{60 with active history} \\
\cmidrule(lr){2-3}\cmidrule(lr){4-5}
& Action agr. $\uparrow$ & Mean Jaccard $\uparrow$ & Action agr. $\uparrow$ & Mean Jaccard $\uparrow$ \\
\midrule
Paraphrase & 90.0\% & 0.920 & 95.0\% & 0.984 \\
Instruction reordering & 96.0\% & 0.973 & 98.3\% & 1.000 \\
\bottomrule
\end{tabular}
\end{adjustbox}
\caption{QTM prompt-sensitivity results. The active-history slice is post hoc and includes inputs where \texttt{NEW\_\allowbreak TOPIC} can trigger archival.}
\label{tab:qtm-prompt-sensitivity-appendix}
\end{table}

The full-sample results show high local agreement under both tested changes, with still higher agreement on the 60 active-history inputs. We additionally report repeat-run, parsing, and candidate-pool checks. On a preselected 30-input repeat set, same-template reruns yield 96.67\%/0.9667 action agreement/Jaccard for the original prompt and 100\%/1.000 for each variant. These repeated runs provide a descriptive reference for within-template variability. All 450 primary calls complete without request errors, production parsing failures, or conservative action fallbacks; three invalid auxiliary \texttt{query\_\allowbreak type} values under instruction reordering are mapped to the standard \texttt{STATIC} default. Effective-action agreement remains close to raw-action agreement (90.0\% versus 88.7\% for paraphrasing and 96.0\% for reordering), while auxiliary schema fields are not perfectly invariant.

High retrieval overlap also persists with larger candidate pools. Historical pools range from 2 to 224 chunks (median 94.5); among the 118 inputs with at least $2k=40$ candidates, mean Jaccard is 0.916 for paraphrasing and 0.975 for instruction reordering. The original prompt produces 127 \texttt{MATCH} and 23 \texttt{NEW\_\allowbreak TOPIC} decisions; paraphrasing changes 15 \texttt{NEW\_\allowbreak TOPIC} decisions to \texttt{MATCH} and none in the opposite direction, while reordering changes five in the former direction and one in the latter. Because the original prompt is not ground truth, these transitions limit claims about subclass invariance but do not establish a quality degradation.

Overall, the probe supports a narrow conclusion: under the tested paraphrase and instruction reordering, QTM largely preserves the effective thread-boundary decision and retrieved memory sources. It does not establish invariance of the complete structured state, accumulated online trajectories, or end-to-end answer quality.

\FloatBarrier
\subsection{QTM Controller Substitution}
\label{app:qtm-controller-swap}

We complement the prompt perturbations with a same-scale controller substitution on the same 150 frozen, stratified reconstructed snapshots. The default Qwen2.5-7B-Instruct controller is replaced by Mistral-7B-Instruct-v0.3 while the original QTM prompt and schema, session histories, historical memory pools, BGE embeddings, production parser, deterministic release logic, and retrieval policy are held fixed. The comparison therefore isolates the controller model within this frozen-snapshot diagnostic; it does not regenerate assistant answers or rerun the online routing trajectory.

We use the same two downstream quantities as in Appendix~\ref{app:qtm-prompt-sensitivity}: the effective \texttt{MATCH}/\allowbreak\texttt{NEW\_\allowbreak TOPIC} decision and Jaccard overlap of the pre-packing top-20 historical memory-source IDs. All controller outputs used in this probe are parseable JSON, and the runs complete without request errors, runtime errors, or parser fallback.

\begin{table}[htbp]
\centering
\footnotesize
\setlength{\tabcolsep}{4pt}
\renewcommand{\arraystretch}{1.08}
\begin{adjustbox}{max width=\linewidth}
\begin{tabular}{@{}lcc@{}}
\toprule
Metric & All 150 inputs & 60 with active history \\
\midrule
Action agreement & 124/150 (82.7\%) & 53/60 (88.3\%) \\
Wilson 95\% interval & 75.8--87.9\% & 77.8--94.2\% \\
Mean retrieval Jaccard & 0.875 & 0.960 \\
Median retrieval Jaccard & 1.000 & 1.000 \\
Identical retrieval set & 115/150 (76.7\%) & 52/60 (86.7\%) \\
Jaccard $\geq 0.9$ & 125/150 (83.3\%) & 55/60 (91.7\%) \\
\bottomrule
\end{tabular}
\end{adjustbox}
\caption{QTM controller-substitution diagnostic. Mistral-7B-Instruct-v0.3 is compared with the Qwen2.5-7B-Instruct original-prompt reference on the same fixed snapshots. The active-history slice is the same 60-input slice used in Table~\ref{tab:qtm-prompt-sensitivity-appendix}.}
\label{tab:qtm-controller-swap}
\end{table}

The two controllers agree on most effective thread-boundary decisions, and the induced retrieval-source sets overlap strongly. The agreement is higher on inputs with active thread history, where a \texttt{NEW\_\allowbreak TOPIC} decision can change archival behavior: action agreement reaches 88.3\%, mean Jaccard reaches 0.960, and 86.7\% of retrieval sets are identical. These results provide controller-substitution evidence at the decision-interface level for one alternative 7B controller, complementing the prompt-sensitivity probe above.

\section{Qualitative and Behavioral Analysis}
\label{app:qualitative}
\label{app:additional_analysis}

\FloatBarrier
\subsection{Context and Memory Failure Cases}
\label{app:failure_cases}

This section provides qualitative examples showing how systems fail under long histories, topic interference, or version-state pressure. The goal is to make the mechanism behind the aggregate metrics visible. The examples below are anonymized from fixed \mc final evaluation outputs where \method passes and multiple fixed full-context baselines fail.

\begin{table}[htbp]
\centering
\small
\setlength{\tabcolsep}{4pt}
\renewcommand{\arraystretch}{1.08}
\begin{tabularx}{\textwidth}{@{}P{0.16\linewidth}YYYP{0.16\linewidth}@{}}
\toprule
Query / setting & Required memory & Baseline pattern & \method pattern & Error type \\
\midrule
Present-tense continuation constraint & Preserve a previously imposed grammar/style rule across turns. & Several full-context baselines lose the active rule. & Maintains the constraint and returns a passing final verdict. & Instruction retention \\
School memory recall under distractors & Recall that a schoolyard feature was located near a sanitation hazard. & Baselines deny the specific detail or lose path continuity. & Recovers the relevant detail and answers consistently. & Inference memory \\
Three-paragraph revision limit & Preserve a local document-format constraint while editing a later version. & Baselines drift on the paragraph-count requirement. & Keeps the active artifact state and satisfies the editing constraint. & Reliable version editing \\
\bottomrule
\end{tabularx}
\caption{Representative anonymized \mc patterns summarized from fixed final-evaluation outputs. They are illustrative rather than additional quantitative evidence.}
\label{tab:failure-cases}
\end{table}

\FloatBarrier
\subsection{Static-Configuration Error Examples on \realm}
\label{app:static_failure}

The following paired examples illustrate missing, ignored, or stale memory but do not establish cause. Aggregate cheap full-context scores are dominated by the overflows in Table~\ref{tab:realmem-full}, so they are not clean evidence of inferior model reasoning.

\begin{table}[htbp]
\centering
\small
\setlength{\tabcolsep}{4pt}
\renewcommand{\arraystretch}{1.08}
\begin{tabularx}{\textwidth}{@{}P{0.16\linewidth}YYYP{0.16\linewidth}@{}}
\toprule
Failure case & Relevant memory evidence & Static-system behavior & \method behavior & Diagnosis \\
\midrule
Travel-plan follow-up after frustration & July travel windows, five adult travelers, and comfort-focused flight constraints. & Gives generic or conflicting travel guidance. & Recovers the travel constraints and responds in the correct planning frame. & Relevant-memory miss \\
Broken workout-video link & A recurring 9 PM back-care routine and selected exercise playlist. & Gives generic troubleshooting or ignores the routine context. & Uses the routine and exercise context to provide a targeted next step. & Context ignored \\
Story-development continuation & The ``Golden Hook'' is locked in and the next component is character development. & Pivots to a generic plot-structure step. & Continues from the locked story state and focuses on the intended component. & Version-state drift \\
\bottomrule
\end{tabularx}
\caption{Representative anonymized \realm static-configuration error patterns; diagnoses are illustrative and descriptive, not causal.}
\label{tab:static-failure-examples}
\end{table}

\FloatBarrier
\subsection{Routing Behavior by Session State}
\label{app:routing_behavior}

This diagnostic summarizes whether the router's behavior changes with QTM state. We summarize model choice and turn cost by realized retrieval scope, interaction phase, and memory-token exposure. Memory-scope slices use \texttt{retrieval\_\allowbreak scope}; phase slices use \texttt{qtm\_\allowbreak phase}. Cheap models group Llama-3.2-3B and Mixtral-8x7B; the mid model is Qwen-72B; the strong model is GPT-4o. High context difficulty is the top third of turns by \texttt{mem\_\allowbreak tokens}. These slices are descriptive diagnostics rather than causal routing rules: model choice depends on the full feature vector and online reward history, so a single QTM field such as \texttt{retrieval\_\allowbreak scope} need not monotonically increase strong-model usage.

\begin{table}[htbp]
\centering
\small
\setlength{\tabcolsep}{5pt}
\renewcommand{\arraystretch}{1.06}
\begin{adjustbox}{max width=\linewidth}
\begin{tabular}{@{}lrrrr@{}}
\toprule
QTM state slice & Cheap model \% & Mid model \% & Strong model \% & Avg. cost \\
\midrule
\texttt{retrieval\_\allowbreak scope=LOCAL} & 12.9 & 62.9 & 24.3 & 0.0199 \\
\texttt{retrieval\_\allowbreak scope=GLOBAL} & 16.6 & 65.6 & 17.8 & 0.0155 \\
\texttt{qtm\_\allowbreak phase=REVISE} & 7.6 & 68.4 & 24.0 & 0.0202 \\
\texttt{qtm\_\allowbreak phase=SYNTHESIZE/VERIFY} & 22.1 & 51.6 & 26.2 & 0.0207 \\
High context difficulty & 14.4 & 62.9 & 22.7 & 0.0206 \\
\bottomrule
\end{tabular}
\end{adjustbox}
\caption{Routing-behavior diagnostic over 1,415 \realm turns. Percentages are within each slice; average cost includes generator and QTM/controller cost when logged.}
\label{tab:routing-behavior}
\end{table}

\section{Deployment Feedback and Reproducibility}
\label{app:deployment_repro}

\FloatBarrier
\subsection{Replacing Turn-Level Judges with Deployment Feedback}
\label{app:deployment_feedback}

The experiments use a turn-level LLM judge to provide scalar feedback for online routing. This choice makes benchmark evaluation reproducible, but the \method interface does not require the reward to come from an LLM judge. The online router only requires a scalar reward associated with the current state, selected model, response, and cost.

In a deployed assistant, this reward could be replaced or supplemented by human or task-derived feedback. Examples include explicit user ratings, edit acceptance, task completion signals, human preference comparisons, workflow success checks, or organization-specific quality audits. Over repeated interactions, such feedback could train a routing policy that reflects user-specific or domain-specific preferences. For example, one user may prefer cheaper models unless quality drops sharply, while another may prefer stronger models for revision or finalization tasks.

This interface allows the routing policy to incorporate application-specific feedback without changing the shared-state memory construction and model-selection pipeline.

\FloatBarrier
\subsection{Reproducibility Checklist}
\label{app:reproducibility}

Table~\ref{tab:reproducibility-checklist} summarizes the experimental setup. Following submission, we will release an implementation package with code, configurations, prompt templates, and evaluation scripts to support reproduction.

\paragraph{Artifact license and intended use.}
Our release is intended to support reproduction of the reported experiments.
It will include code, prompt templates, configurations, and evaluation scripts, but it will not redistribute benchmark data or private intermediate logs.
Users should obtain \realm and \mc from their original sources and follow their respective licenses and terms.
We will release our implementation artifacts under the Apache License 2.0.

\begin{table}[htbp]
\centering
\small
\setlength{\tabcolsep}{4pt}
\renewcommand{\arraystretch}{1.10}
\begin{tabularx}{\textwidth}{@{}P{0.25\linewidth}Y@{}}
\toprule
Item & Reproducibility information \\
\midrule
Code and configs & An implementation package will be released with system configurations, prompt templates, and evaluation scripts. \\
\realm split & \realm evaluation pool with 10 personas and 1,415 queries; QA is normalized as official raw QA divided by 3. \\
\mc split & Converted \mc split with 245 scenarios and 1,480 turns; chunk size 3; target subset consists of \textsc{Inference Memory} and \textsc{Reliable Version Editing}. \\
Ablation slice & Three-persona \realm evaluation slice with 473 turns (472 valid QA evaluations for the full variant); personas sampled with seed 28; matched top-10 historical retrieval budget. \\
Candidate model pools & Main \realm runs use the configured \method candidate pool reported in Appendix~\ref{app:router_features}; ablations use Llama3B, Qwen2.5 72B, and GPT-4o. \\
QTM and router settings & Main QTM uses \texttt{qwen/qwen-2.5-7b-instruct}; its prompt, schema, guardrails, and state-to-memory mapping are fixed. The controller-swap diagnostic additionally evaluates Mistral-7B-Instruct-v0.3 with the same original prompt/schema. Only the router updates online in the main system. \\
Feedback and evaluators & GPT-4o-mini supplies online routing feedback; primary benchmark final evaluators are described in Appendix~\ref{app:final_judges}. The interaction-order diagnostic also uses a fixed GPT-4o-mini path-aware evaluator across all replays. \\
Reward and cost & Reward aggregation is specified in Appendix~\ref{app:turn_judge}; rubric dimensions appear in Table~\ref{tab:turn-judge-rubric}. Main \realm uses $\lambda_{\mathrm{cost}}=0.002$; component ablations use $\lambda_{\mathrm{cost}}=0.5$. Cost accounting follows Appendix~\ref{app:cost_accounting}. \\

Cost reporting & Main \realm tables include generator and controller API costs (Appendix~\ref{app:cost_accounting}); historical \mc costs use the original run-level convention. Final evaluators and local compute are excluded. \\
\bottomrule
\end{tabularx}
\caption{Reproducibility checklist. The release package will provide the implementation, configurations, prompt templates, and evaluation scripts.}
\label{tab:reproducibility-checklist}
\end{table}

\end{document}